\documentclass[journal]{IEEEtran}

\usepackage{cite}
\usepackage{amsmath,amssymb,amsfonts}
\usepackage{graphicx}
\usepackage{textcomp}
\usepackage[colorlinks=true, linkcolor=blue, citecolor=blue, urlcolor=blue]{hyperref}
\usepackage{tabularx}
\usepackage{xcolor}
\usepackage{booktabs}
\usepackage{makecell}
\usepackage{pifont}
\usepackage{multirow}
\usepackage{arydshln}
\newcommand{\cmark}{\ding{51}} 

\definecolor{gainGreen}{RGB}{0,100,0}
\definecolor{gainRed}{RGB}{150,0,0}

\usepackage{bm}          
\usepackage{array}       
\usepackage{arydshln}   

\newcommand{\best}[1]{$\bm{#1}$}
\newcommand{\second}[1]{$\underline{#1}$}
\newcommand{\plain}[1]{$#1$}
\newcommand{\fail}{\ensuremath{\times}}

\newcommand{\gaincell}[2]{%
  \makecell[c]{%
    #1\\[0.5mm]
    {\fontsize{5.5}{7}\selectfont #2}%
  }%
}

\newcommand{\posgain}[1]{%
  \textcolor{gainGreen}{$\uparrow$\,#1\%}%
}

\newcommand{\neggain}[1]{%
  \textcolor{gainRed}{$\downarrow$\,#1\%}%
}

\usepackage{bm}
\makeatletter
\AtBeginDocument{\DeclareMathVersion{bold}
\DeclareSymbolFont{NewLetters}{T1}{times}{m}{it}
\SetSymbolFont{operators}{bold}{T1}{times}{b}{n}
\SetSymbolFont{NewLetters}{bold}{T1}{times}{b}{it}
\SetMathAlphabet{\mathrm}{bold}{T1}{times}{b}{n}
\SetMathAlphabet{\mathit}{bold}{T1}{times}{b}{it}
\SetMathAlphabet{\mathbf}{bold}{T1}{times}{b}{n}
\SetMathAlphabet{\mathtt}{bold}{OT1}{pcr}{b}{n}
\SetSymbolFont{symbols}{bold}{OMS}{cmsy}{b}{n}
\renewcommand\boldmath{\@nomath\boldmath\mathversion{bold}}}
\makeatother

\title{Desc++: Efficient Descriptor Enhancement for Data Association in Existing Visual SLAM Systems}

\author{
Ting-Wei Ou\textsuperscript{1},
Huang-Ting Lin\textsuperscript{2},
and Kuu-Young Young\textsuperscript{2}
\IEEEmembership{Senior Member, IEEE}\\[6pt]
\textit{Human \& Machine Lab, National Yang Ming Chiao Tung University}\\
\textit{Graduate Degree Program of Robotics\textsuperscript{1}}\\
\textit{Institute of Electrical and Control Engineering\textsuperscript{2}}
}

\begin{document}
\maketitle
\thispagestyle{empty}
\pagestyle{empty}

\begin{abstract} 
Reliable visual data association is fundamental to visual SLAM (V-SLAM), as it directly determines the quality of the camera pose estimation and map consistency. However, the handcrafted descriptors used by most mature real-time systems degrade under illumination and viewpoint changes, while learning-based front-ends that address this weakness typically require replacing the extraction-and-matching pipeline and introduce substantial computational overhead. Descriptor enhancement offers a compromise by refining existing descriptors within their original format, yet current methods rely on simplified attention mechanisms whose limited contextual modeling constrains the achievable matching quality. To resolve this trade-off between contextual expressiveness and efficiency, we propose Desc++, a lightweight enhancement module that jointly encodes descriptor representations and keypoint geometry and aggregates spatial context through a hybrid architecture that combines order-agnostic global attention with geometry-aware sequential modeling in linear time. The enhanced descriptors retain their original dimensionality and matching interface, enabling integration into deployed V-SLAM systems without modifying the pipeline. Experiments across descriptor matching, correspondence analysis, and system-level benchmarks with four different V-SLAM systems demonstrate that Desc++ improves matching accuracy over the state-of-the-art enhancement method, translates these gains into more accurate and stable trajectory estimation, and achieves a favorable balance between accuracy and efficiency for practical integration into existing real-time V-SLAM pipelines. The source code and pretrained model weights are available at: \href{https://github.com/ouotingwei/DescPP.git}{https://github.com/ouotingwei/DescPP.git}.
\end{abstract}

\section{Introduction}
Autonomous systems operating in unknown, GPS-denied environments require continuous state estimation from onboard sensing, a capability for which visual SLAM (V-SLAM) has become a core solution in robotics~\cite{b45}. The performance of a V-SLAM system largely depends on reliable visual data association, which establishes feature correspondences across images for camera tracking, pose optimization, and map construction. In sparse feature-based pipelines, these correspondences are established by matching local feature descriptors extracted around detected keypoints. Consequently, descriptor discriminability directly affects matching accuracy and overall SLAM robustness. To meet the computational requirements of real-time applications, most mature V-SLAM systems~\cite{b2, b3, b4, b41} rely on handcrafted descriptors, which are computationally efficient but remain sensitive to illumination changes and viewpoint variations, often leading to degraded data association and accumulated trajectory drift.

To improve visual data association, recent studies have explored learning-based local feature extraction~\cite{b8, b9, b10} and feature matching~\cite{b12, b13, b14}. Although these approaches substantially improve matching robustness and accuracy, integrating them into existing V-SLAM systems generally requires replacing the original front-end pipeline, limiting their adoption in deployed real-time systems. Descriptor enhancement~\cite{b7} provides an alternative design strategy by improving the discriminative capability of existing feature descriptors while preserving the original detector, descriptor format, and matching interface. Such a plug-and-play design enables deployed V-SLAM systems to benefit from learned representations with minimal system modification. Contextual relationships among neighboring keypoints provide complementary geometric and structural cues that are beneficial for descriptor discrimination. However, the state-of-the-art enhancement method~\cite{b7} typically relies on lightweight, simplified attention mechanisms to meet real-time constraints, thereby limiting its ability to capture rich contextual relationships amid challenging appearance changes. This reflects a fundamental trade-off between contextual expressiveness and computational efficiency in descriptor enhancement for V-SLAM.

To address this trade-off, we propose Desc++, a lightweight descriptor enhancement framework that improves descriptor discriminability while preserving the computational efficiency and pipeline compatibility required by real-time V-SLAM. Desc++ jointly encodes descriptor representations and keypoint geometry, and aggregates spatial context through a hybrid architecture that models both global and sequential dependencies in linear time, producing enhanced descriptors that directly replace the original ones without any modification to the downstream pipeline.

This work makes three contributions toward practical descriptor enhancement for deployed V-SLAM systems. First, we design a hybrid context aggregation architecture that operates in parallel: an attention-free branch provides order-agnostic global context, while a Mamba branch models geometry-aware sequential dependencies along a Z-order serialized keypoint sequence. This design captures richer contextual relationships than the simplified attention used in prior enhancement methods~\cite{b7} while retaining linear-time complexity, as validated on the HPatches benchmark~\cite{b31}, where Desc++ improves matching accuracy over the state-of-the-art enhancer across both handcrafted and learned descriptors. Second, we demonstrate that the enhanced descriptors function as a genuine plug-and-play component: the same module is integrated, without modifying any downstream tracking or matching logic, into four heterogeneous systems: a stereo system (ORB-SLAM2~\cite{b2}), a stereo-inertial system (ORB-SLAM3~\cite{b3}), a visual-LiDAR system (RGB-L~\cite{b4}), and a multi-camera visual-inertial system (MAVIS-SLAM~\cite{b41})---improving trajectory accuracy on the majority of sequences across the EuRoC~\cite{b33}, KITTI~\cite{b32}, and Hilti SLAM Challenge 2023~\cite{b40} benchmarks. Third, through a correspondence-level analysis and a comparison with a learned front-end~\cite{b42}, we characterize when and why descriptor enhancement is effective: improved descriptor discriminability yields more tracked map points and tighter geometric constraints, achieving accuracy competitive with replacement-based learned pipelines at a fraction of their computational cost, while remaining bounded by the quality of the initial feature extraction. Overall, these results indicate that refining descriptor quality is a practical and efficient path to improving existing real-time V-SLAM systems without redesigning their front-ends.


\section{Related Work}
This section reviews representative techniques related to feature-based visual data association, including local features, learning-based data association, and efficient visual context modeling.

\subsection{Local Features For Visual SLAM}
Local features typically consist of two components: a detector that identifies salient keypoints in an image and a descriptor that encodes the local appearance of these keypoints into compact representations. Traditional approaches rely on handcrafted designs with distinct trade-offs. SIFT~\cite{b22} uses gradient histograms to achieve scale- and rotation-invariance, prioritizing robustness for Structure-from-Motion (SfM)~\cite{b37}. In contrast, ORB~\cite{b1} prioritizes computational efficiency by combining FAST detection~\cite{b24} with binary intensity comparisons~\cite{b23}, making it suitable for real-time applications. However, both remain inherently constrained by their reliance on fixed prior knowledge. Learning-based local features have emerged as an alternative to handcrafted ones~\cite{b8, b9, b10}, employing Convolutional Neural Networks (CNNs) within a "detect-and-describe" paradigm that jointly optimizes the detector and descriptor. By learning directly from large-scale data, these approaches overcome the limitations of fixed heuristics, yielding representations that are significantly more discriminative and robust under challenging conditions. Recent works have demonstrated the effectiveness of integrating learning-based features into existing SLAM systems. GCNv2-SLAM~\cite{b39} redesigns the original GCN~\cite{b44} into a lightweight single-frame architecture with binary descriptors compatible with ORB~\cite{b1}, enabling real-time deployment in ORB-SLAM2~\cite{b2}. In contrast, LF2SLAM~\cite{b38} replaces the ORB keypoint detector with a task-specific SuperPoint~\cite{b8} network while retaining the binary descriptors and the downstream SLAM pipeline. These studies suggest that improving feature quality can translate into more accurate and robust SLAM performance.

\begin{figure*}[t] 
    \centering
    \includegraphics[width=\textwidth]{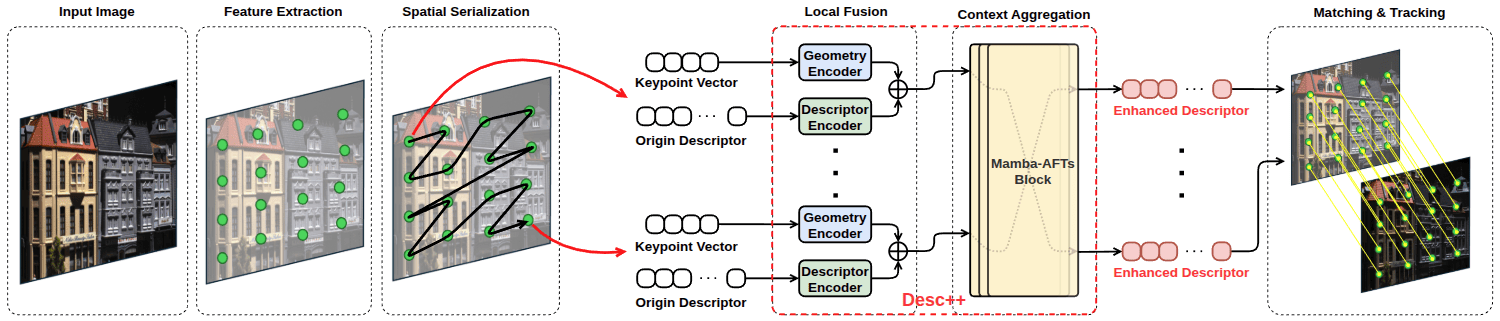} 
    \caption{The overall pipeline of \textbf{Desc++}. First, unstructured keypoints extracted from an image are spatially organized via Z-ordering. In the Local Fusion Stage, the geometry and original descriptor of each keypoint are fused to enrich the local representation with geometric priors. These features are then processed by the Context Aggregation Stage, where the hybrid Mamba-AFTs Block performs relation-aware modeling to capture interdependencies among features. Finally, the enhanced descriptors are used for robust matching and tracking in downstream V-SLAM tasks.}
    \label{fig:Fig1} 
\end{figure*}

\subsection{Learning-based Data Association}
Learning-based approaches have become an effective strategy for improving visual data association by incorporating contextual information beyond descriptor similarity. Existing methods can be categorized into learned matching and descriptor enhancement, which differ in their integration strategy and deployment characteristics.

\subsubsection{Learned Matching}
Traditional feature matching relies on nearest-neighbor search followed by geometric verification, such as Lowe's ratio test~\cite{b22} and RANSAC~\cite{b21}. Although computationally efficient, these methods only exploit descriptor similarity and ignore contextual relationships among keypoints. Learned matching methods, including SuperGlue~\cite{b12}, LightGlue~\cite{b13}, and MambaGlue~\cite{b14}, address this limitation by jointly reasoning over feature correspondences through graph neural networks, attention mechanisms, or hybrid models. These approaches substantially improve the robustness of matching under challenging conditions. However, they typically replace the original matching module and require additional feature interaction during inference, making integration into mature real-time V-SLAM systems more computationally demanding.

\subsubsection{Descriptor Enhancement}

Descriptor enhancement follows a different strategy, refining existing descriptors while preserving the original feature representation and the matching interface. Early work~\cite{b36} projected handcrafted descriptors into more discriminative embedding spaces, followed by context-aware refinement methods that introduced contextual aggregation to improve descriptor quality. ContextDesc~\cite{b11} aggregates cross-modality context by combining geometric cues from keypoint distribution with visual context derived from regional image representations extracted by an off-the-shelf retrieval backbone. While effective, this design requires access to the raw image and an additional deep backbone at inference time, which conflicts with the descriptor-level, image-free interface required for drop-in integration into deployed V-SLAM front-ends, and its evaluation is confined to floating-point descriptors. FeatureBooster~\cite{b7}, in contrast, operates purely on the extracted descriptors and their keypoint geometry, supports binary descriptors, and is therefore the directly comparable prior work under our deployment setting. Unlike learned matching methods that jointly optimize feature correspondence during inference, descriptor enhancement improves data association by refining descriptors while retaining the original matching pipeline. However, existing approaches still face a trade-off between contextual modeling capability and computational efficiency, motivating the development of more effective lightweight context aggregation architectures.

\subsection{Efficient Visual Context Modeling}

Transformer-based architectures have demonstrated remarkable capability in modeling long-range visual context and have been widely adopted in vision tasks ranging from image understanding to feature matching~\cite{b12,b13,b15,b25,b26}. However, their quadratic computational complexity and large Key-Value memory footprint limit their applicability to real-time and resource-constrained systems.

To address these, Mamba~\cite{b17} leverages a Selective Scan Mechanism to model long-range dependencies with linear complexity. Meanwhile, the inherent autoregressive nature of State Space Models (SSMs) poses challenges for processing visual data, where spatial relationships are noncausal. Pure Mamba-based approaches often struggle to capture a comprehensive global context in a single forward pass, necessitating multi-directional scanning strategies (e.g., bidirectional~\cite{b18} or cross-scan~\cite{b27}) to restore spatial integrity. However, these strategies inevitably introduce additional latency and computational overhead. Consequently, SSM-based visual backbones are largely distinguished by how they reconcile the causal nature of SSMs with the noncausal structure of image data, whether through repeated scans over a regular lattice~\cite{b18,b27} or through hybridization with attention~\cite{b19}. Desc++ shares the motivation of linear-time global context, but operates on a sparse, unordered set of keypoints rather than a dense pixel grid, and therefore resolves the ordering problem by an explicit spatial serialization of the keypoint set rather than by repeated scanning.

Recent hybrid architectures~\cite{b14,b19} attempt to mitigate the limitations of pure SSMs by combining them with attention mechanisms. Among these, MambaGlue~\cite{b14} is the most closely related instance within the feature-association domain, but it differs from the present work in both the pipeline stage it targets and the way the two mechanisms are combined. In terms of stage, MambaGlue operates at the matching stage, consuming descriptors from an image pair and jointly reasoning over their cross-image correspondences, thereby replacing the matcher and requiring pairwise feature interaction at inference. Desc++ instead operates at the descriptor stage, refining the descriptors of a single frame in one feed-forward pass so that the original nearest-neighbor matcher and its distance metric remain unchanged. In terms of composition, coupling Mamba with standard Transformer attention in series reintroduces quadratic complexity, thereby undermining the linear-time advantages intrinsic to SSMs. In contrast, Desc++ avoids both the latency of multi-directional scanning and the quadratic cost of serial Mamba--attention coupling: it employs a single Z-order serialization to preserve spatial locality for the SSM branch, and pairs it in parallel with an attention-free transformer branch~\cite{b16} that supplies an order-agnostic global receptive field, compensating for the directional bias of sequential scanning while keeping the overall complexity linear.


\section{The Desc++ Architecture}
In this section, we introduce Desc++, a lightweight descriptor
enhancement module designed to balance matching robustness with
low-latency inference on embedded platforms. We first formulate the descriptor enhancement problem, then detail the two-stage architecture consisting of a Local Fusion stage and a Context Aggregation stage, followed by the training objective and implementation details.

\subsection{Problem Formulation}
Given an input image $I$, a feature detector extracts $N$ keypoints, represented by their geometries $\mathcal{K} = \{k_i \in \mathbb{R}^{M}\}_{i=1}^{N}$ and corresponding existing descriptors $\mathcal{D} = \{d_i \in \mathbb{R}^{D}\}_{i=1}^{N}$. Specifically, $k_i$ captures $M$ geometric attributes, while $d_i$ is a $D$-dimensional feature vector. The goal of descriptor enhancement is to learn a lightweight mapping $f_{\theta}$:

\begin{equation}
    f_{\theta} : (\mathcal{D}, \mathcal{K}) \rightarrow \hat{\mathcal{D}},
    \label{eq:enhancement_mapping}
\end{equation}

where $\hat{\mathcal{D}} = \{\hat{d}_i \in \mathbb{R}^{D}\}_{i=1}^{N}$ is the set of enhanced descriptors. Note that $\hat{d}_i$ retains the same dimensionality $D$ as the input, but incorporates global/local context information.
The goal is to let $\hat{\mathcal{D}}$ exhibit higher discriminability and stability under illumination or viewpoint changes. 
In SLAM or matching tasks, $\hat{\mathcal{D}}$ can directly replace $\mathcal{D}$, ensuring plug-and-play compatibility. Throughout the paper, the subscript $i$ indexes the $N$ keypoints of a frame, whereas the subscript $t$, used in Eq.~\ref{eq:mamba}, indexes positions in the Z-order-serialized keypoint sequence.

To realize $f_{\theta}$, we propose Desc++, a lightweight module designed to enhance existing feature descriptors. The architecture follows a two-stage enhancement strategy: a Local Fusion stage that integrates descriptor and geometric representations, and a Context Aggregation stage that captures inter-keypoint relations through the proposed Mamba-AFTs Block, shown in Fig.~\ref{fig:Fig1}, which illustrates the overall pipeline of the proposed descriptor enhancement process.

\subsection{Z-order Spatial Serialization}
The extracted feature sets $\mathcal{K}$ and $\mathcal{D}$ are initially unstructured collections that lack the sequential organization mandated by the causal nature of SSMs. To address this, we restructure the input using Z-order serialization~\cite{b28}, sorting the keypoints by Morton codes derived from their pixel coordinates. This mapping projects the 2D layout onto a 1D sequence, ensuring that geometric neighbors appear in close proximity in the index. Consequently, the serialized input preserves strong local correlations, enabling the subsequent Mamba block to efficiently aggregate spatial context.

\subsection{Local Fusion Stage}
For each keypoint $k_i$ detected in the image with its existing descriptor $d_i$, the Local Fusion stage fuses the descriptor and its keypoint geometry to generate a geometry-aware embedding.

\noindent \textbf{Descriptor Encoder}:
To refine the off-the-shelf descriptor $d_i$, we employ an MLP with a residual connection: the MLP performs a non-linear transformation that projects the detector-provided features into a learnable latent space, while the residual path
preserves the original descriptor information, yielding the embedding $\mathbf{e}^{\text{desc}}_i$:

\begin{equation}
    \label{eq:desc-enc}
    \mathbf{e}^{\text{desc}}_i = d_i + \mathrm{MLP}_{\text{desc}}(d_i).
\end{equation}

\noindent \textbf{Geometric Encoder}:
To encode the keypoint geometry $k_i \in \mathbb{R}^{M}$, we map it to a feature embedding via Learnable Fourier Features (LFF)~\cite{b20}. The geometric attributes are first projected into a frequency-based representation through a
sinusoidal mapping $\Phi_{\text{LFF}}(\cdot)$, and an MLP then modulates the resulting features to obtain the geometric representation $\mathbf{e}^{\text{geo}}_i$:

\begin{equation}
    \label{eq:geo-enc}
    \mathbf{e}^{\text{geo}}_i = \mathrm{MLP}_{\text{geo}}\big(\Phi_{\text{LFF}}(k_i)\big).
\end{equation}

Since the geometric attributes are heterogeneous in physical dimension and dynamic range, $\Phi_{\text{LFF}}(\cdot)$ operates on a partition of $k_i$ into $G_p$ groups $\{\mathbf{p}_{i,g} \in \mathbb{R}^{m_g}\}_{g=1}^{G_p}$ with
$\sum_{g} m_g = M$. For detectors that provide scale and orientation (e.g., ORB and SIFT), the groups correspond to pixel coordinates, scale, and orientation ($m_g = 2, 1, 1$); for detectors without explicit scale or orientation, the
available attributes are used instead, as listed in Table~\ref{tab:model_details}. Each group is projected by a learnable frequency matrix $\mathbf{W}_g \in \mathbb{R}^{(F_g/2)\times m_g}$, initialized from $\mathcal{N}(0, \sigma_g^{2})$ with a group-specific $\sigma_g$, yielding the Fourier embedding

\begin{equation}
    \label{eq:lff-group}
    \boldsymbol{\phi}_g(\mathbf{p}_{i,g}) = \tfrac{1}{\sqrt{F_g}}
    \big[\cos(\mathbf{W}_g\,\mathbf{p}_{i,g});\ \sin(\mathbf{W}_g\,\mathbf{p}_{i,g})\big]
    \in \mathbb{R}^{F_g},
\end{equation}
so that $\Phi_{\text{LFF}}(k_i) = [\boldsymbol{\phi}_1; \dots;
\boldsymbol{\phi}_{G_p}] \in \mathbb{R}^{\sum_g F_g}$.

Assigning a separate $\mathbf{W}_g$ per group follows from the structure of the input rather than from tuning: the attributes differ in dimensionality and in physical units, so a shared frequency matrix would require padding them to a
common dimension and would impose a single bandwidth on quantities that are not commensurable. The partition therefore confines the shift-invariant inductive bias of LFF to within each commensurable group, while cross-attribute
interactions are delegated to $\mathrm{MLP}_{\text{geo}}$. Pixel coordinates are normalized before encoding, whereas the remaining attributes are used in their native units. The values of $\sigma_g$ were selected empirically; since
$\mathbf{W}_g$ is learnable, they set the initial frequency band rather than a fixed range. The complete configuration, including $F_g$, $\sigma_g$, and the MLP dimensions for each detector, is listed in Table~\ref{tab:model_details}.

\noindent The geometric and descriptor embeddings are then fused via element-wise summation to obtain the intermediate fused feature $\mathbf{z}_i$:

\begin{equation}
    \label{eq:fuse}
    \mathbf{z}_i = \mathbf{e}^{\text{desc}}_i + \mathbf{e}^{\text{geo}}_i.
\end{equation}

\begin{figure}[t] 
    \centering
    \includegraphics[width=0.48\textwidth]{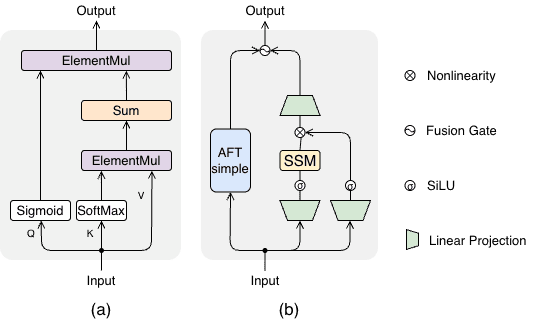} 
    \caption{Detailed diagrams of (a) AFT-Simple \cite{b16} vs. (b) Mamba-AFTs Block.} 
    \label{fig:Fig2} 
\end{figure}

\subsection{Context Aggregation Stage}
Taking the fused features $\mathcal{Z} = \{\mathbf{z}_i\}_{i=1}^{N}$ as input, we design the Mamba-AFTs Block to model contextual dependencies with minimal overhead. As illustrated in Fig.~\ref{fig:Fig2}, the block operates in two parallel branches to leverage the complementary strengths of sequence-sensitive state-space models and order-agnostic attention.

\begin{table*}[t]
\centering
\caption{Detailed architectural specifications of Desc++ for each feature type. $D$ denotes the descriptor dimension. MLPs use ReLU activations between hidden layers. The LFF frequency matrices $\mathbf{W}_g \in \mathbb{R}^{(F_g/2) \times m_g}$ are learnable and initialized from $\mathcal{N}(0, \sigma_g^{2})$, as defined in Sec.~III-C. For ORB and SIFT the three groups are pixel coordinates, scale, and orientation; for SuperPoint and ALIKE they are pixel coordinates and the detection score.}
\label{tab:model_details}
\resizebox{\textwidth}{!}{
\begin{tabular}{llcccc}
\toprule
\textbf{Module} & \textbf{Configuration} & \textbf{ORB~\cite{b1}} & \textbf{SIFT~\cite{b22}} & \textbf{SuperPoint~\cite{b8}} & \textbf{ALIKE~\cite{b9}} \\ 
\midrule
General 
 & Descriptor dim. $D$ & 256 & 128 & 256 & 128 \\
\midrule
\multirow{4}{*}{LFF Geometric Enc.} 
 & Input groups $(m_1, m_2, \dots)$ & $(2,1,1)$ & $(2,1,1)$ & $(2,1)$ & $(2,1)$ \\
 & Fourier dims $(F_1, F_2, \dots)$ & $(64,32,32)$ & $(64,32,32)$ & $(64,64)$ & $(64,64)$ \\
 & Init. scales $(\sigma_1, \sigma_2, \dots)$ & $(2.5{\times}10^{-3},\,4{\times}10^{-2},\,1)$ & $(2.5{\times}10^{-3},\,4{\times}10^{-2},\,1)$ & $(2.5{\times}10^{-3},\,4{\times}10^{-2})$ & $(2.5{\times}10^{-3},\,4{\times}10^{-2})$ \\
 & $\mathrm{MLP}_{\text{geo}}$ & $128{-}128{-}256{-}256$ & $128{-}128{-}128{-}128$ & $128{-}128{-}256{-}256$ & $128{-}128{-}128{-}128$ \\
\midrule
Descriptor Enc.
 & $\mathrm{MLP}_{\text{desc}}$ (residual) & $256{-}512{-}256{-}256$ & $128{-}256{-}128{-}128$ & $256{-}512{-}256{-}256$ & $128{-}256{-}128{-}128$ \\
\midrule
\multirow{5}{*}{Mamba-AFTs Block} 
 & \# CA layers $N_{\text{CA}}$ & 2 & 2 & 2 & 6 \\
 & SSM state dim. $d_{\text{state}}$ & 16 & 16 & 16 & 16 \\
 & Conv. kernel size & 3 & 3 & 3 & 3 \\
 & Expansion $E$ (inner dim. $ED$) & 2 (512) & 2 (256) & 2 (512) & 2 (256) \\
 & $\Delta$-rank $\lceil D/16 \rceil$ & 16 & 8 & 16 & 8 \\
\midrule
FFN
 & MLP ($D{-}2D{-}D$) & $256{-}512{-}256$ & $128{-}256{-}128$ & $256{-}512{-}256$ & $128{-}256{-}128$ \\
\midrule
\multirow{3}{*}{Output Head} 
 & Final projection & $256{-}256$ & $128{-}128$ & $256{-}256$ & $128{-}128$ \\
 & Activation & Tanh & -- & -- & -- \\
 & $\ell_2$-normalize & \ding{55} & \ding{51} & \ding{51} & \ding{51} \\
\midrule
Regularization
 & Dropout $p$ & 0.1 & 0.1 & 0.1 & 0.1 \\
\bottomrule
\end{tabular}
}
\end{table*}

\noindent\textbf{AFT-Simple Branch:}
To capture global context without the path dependency of SSMs or the quadratic bottleneck of standard Transformers, we employ the AFT-Simple mechanism~\cite{b16}. Unlike Multi-Head Attention, AFT decouples the interaction between queries and keys. It serves as a permutation-invariant global memory that allows each keypoint to instantaneously incorporate context from the entire feature set. To formalize this process, the global context is computed through a weighted aggregation of the key-value pairs as follows.

Given the input features $\mathbf{X} \in \mathbb{R}^{N \times D}$, we project them into queries $\mathbf{Q}$, keys $\mathbf{K}$, and values $\mathbf{V}$. The operation for the $i$-th feature index $\mathbf{Y}_{\text{AFT}, i}$ is described as
\begin{equation}
\label{eq:aft}
\mathbf{Y}_{\text{AFT}, i} = \sigma(\mathbf{Q}_i) \odot \frac{\sum_{j=1}^{N} \exp(\mathbf{K}_j) \odot \mathbf{V}_j}{\sum_{j=1}^{N} \exp(\mathbf{K}_j)},
\end{equation}
where $N$ is the number of keypoints (the sequence length after serialization), $\sigma(\cdot)$ is the sigmoid gate, and $\odot$ represents element-wise multiplication. This formulation allows each query to interact with the global context in $\mathcal{O}(ND)$ time complexity, providing a global context that complements the local focus of the Mamba branch.

\noindent\textbf{Mamba Branch:}
To efficiently capture local and structural dependencies along the serialized path, we utilize a Mamba block~\cite{b17}. Its core lies in a selective scan mechanism, in which the state-transition parameters $(\bar{\mathbf{A}}_t, \bar{\mathbf{B}}_t)$ become input-dependent, where $\bar{\mathbf{A}}_t$ and $\bar{\mathbf{B}}_t$ denote the discretized state transition and input matrices at step $t$, respectively. This allows the model to selectively propagate geometric priors or reset context based on the current input. The underlying recurrence for the $t$-th step in the sequence is described as

\begin{equation}
\label{eq:mamba}
\begin{split}
    &\mathbf{h}_t = \bar{\mathbf{A}}_t \mathbf{h}_{t-1} + \bar{\mathbf{B}}_t \mathbf{x}_t \\
    &\mathbf{y}_t = \mathbf{C}_t \mathbf{h}_t,
\end{split}
\end{equation}

where $\mathbf{x}_t$ denotes the $t$-th element of the Z-order-serialized feature sequence, $\mathbf{h}_t$ is the hidden state, and $\bar{\mathbf{A}}_t, \bar{\mathbf{B}}_t, \mathbf{C}_t$ are derived from $\mathbf{x}_t$. This content-aware recurrence enables precise modeling of the geometric continuity preserved by the Z-order serialization. Since the serialization defines a bijection between sequence positions and keypoints, we hereafter re-index the Mamba-branch output by keypoint and denote it as $\mathbf{Y}_{\text{Mamba},i}$ for keypoint $i$. The two branches are then fused through a lightweight learned fusion mechanism. Specifically, for each keypoint $i$, the outputs of the two branches are concatenated and projected by a learned linear layer into two logits, which a softmax normalizes into a pair of complementary scalar weights $w_{\text{AFT},i}$ and $w_{\text{Mamba},i}$ with $w_{\text{AFT},i} + w_{\text{Mamba},i} = 1$. The fused output of the block is then obtained as their convex combination:

\begin{equation}
\mathbf{Y}_{\text{out},i} = w_{\text{AFT},i}\,\mathbf{Y}_{\text{AFT},i}
+ w_{\text{Mamba},i}\,\mathbf{Y}_{\text{Mamba},i}.
\end{equation}

This gating lets the network learn, independently for each layer, how to combine the order-agnostic global context from the AFT branch with the sequence-sensitive local geometric dependencies captured by the Mamba branch, rather than imposing a fixed combination a priori. As analyzed in Sec.~\ref{sec:gate_analysis}, the two layers learn distinct fusion regimes, with the first specializing almost entirely to the Mamba branch and the second maintaining a stable mixture of the two. Each Context Aggregation layer wraps the Mamba-AFTs block with a residual connection, layer normalization, and a position-wise feed-forward network. We stack $N_{\text{CA}}$ such layers; the complete per-format configuration of all modules is listed in Table~\ref{tab:model_details}. The output of the final Context Aggregation layer, after a linear projection and the format-specific activation described in the Training Details, constitutes the enhanced descriptor $\hat{d}_i$ defined in Eq.~(\ref{eq:enhancement_mapping}).

\subsection{Loss Function}
As in previous work \cite{b7}, we treat the descriptor matching task as a nearest-neighbor retrieval problem and design the training objective accordingly. It is guided by a composite loss based on FastAP~\cite{b29}, a differentiable implementation of the Average Precision (AP) metric. The total loss, $\mathcal{L}_{\text{total}}$, consists of two components: a primary matching loss ($\mathcal{L}_{\text{match}}$) and a booster loss ($\mathcal{L}_{\text{boost}}$). The former aims to enhance the discriminative capability of the refined descriptors $\hat{d}$, while the latter encourages consistent improvement over the original descriptors $d$. The overall objective is expressed as the sum of the two terms:

\begin{equation}
\label{eq:loss_total_condensed}
\mathcal{L}_{\text{total}} = \underbrace{\frac{1}{N} \sum_{i=1}^{N} (1 - \text{AP}(\hat{d}_i))}_{\mathcal{L}_{\text{match}}} + \underbrace{\frac{1}{N} \sum_{i=1}^{N} \max\left(0, \frac{\text{AP}(d_i)}{\text{AP}(\hat{d}_i)} - 1\right)}_{\mathcal{L}_{\text{boost}}}.
\end{equation}

Ground-truth correspondences are obtained by warping keypoints across views using the provided depth and camera poses. For each anchor descriptor, keypoints within a $3$-pixel radius of the warped location are treated as positives and those beyond $16$ pixels as negatives, while the intermediate region is excluded from supervision to avoid label ambiguity; anchors without any positive are discarded, and the loss is accumulated over both matching directions of each image pair. Regarding numerical stability, AP is computed via a soft-histogram approximation with $B = 10$ distance bins and guarded normalization, and since every retained anchor has at least one positive falling into a valid bin, the per-descriptor AP is strictly positive. Consequently, the denominator $\text{AP}(\hat{d}_i)$ in $\mathcal{L}_{\text{boost}}$ is bounded away from zero, and the ratio remains well-defined throughout training.

\subsection{Training Details}
\label{sec:train}
\noindent\textbf{Training Process: } Desc++ was trained on the MegaDepth dataset~\cite{b30}, following a training protocol similar to FeatureBooster~\cite{b7} and utilizing the same training scenes as DISK~\cite{b10}. For each scene, 300 image pairs were randomly sampled, with overlapping scores ranging from 0.1 to 1, as determined by D2-Net~\cite{b35}. From each image pair, corresponding $512\times512$ patches were cropped, and up to 2048 features per image were extracted. Specifically, we employed the ORB-SLAM's extractor~\cite{b2} for ORB~\cite{b1} features, the COLMAP framework~\cite{b6, b37} for SIFT~\cite{b22} features, and official pre-trained weights for SuperPoint~\cite{b8} and ALIKE-L~\cite{b9}. The training objective was to optimize the ranking capability of the enhanced descriptors by maximizing the Average Precision (AP) for geometrically consistent ground-truth matches.

\noindent\textbf{Model Details: } Table~\ref{tab:model_details} summarizes the architectural configurations tailored for each feature type. Regarding keypoint geometry, we partition pixel coordinates and other geometric attributes into independent groups, which are then projected into a 128-dimensional latent space using Learnable Fourier Features~\cite{b20}. For the Context Aggregation stage, we employ two layers for ORB~\cite{b1}, SIFT~\cite{b22}, and SuperPoint~\cite{b8}. For ALIKE~\cite{b9}, the number of layers is increased to six to maintain a model capacity comparable to FeatureBooster~\cite{b7}, ensuring a fair comparison rather than introducing additional capacity. To maintain compatibility with the Hamming-distance matching used by binary descriptors, the enhanced ORB descriptors are constrained by a Tanh activation and binarized to $\{-1,+1\}$ via a sign function.
Crucially, this binarization is applied during training as well, using a straight-through estimator (STE): the forward pass operates on the quantized descriptors, so the FastAP loss is computed directly on Hamming distances, while gradients bypass the non-differentiable sign operation and flow through the Tanh-constrained outputs. This ensures that the training objective and the inference-time matching metric are identical, eliminating any quantization-induced train--test mismatch. The Tanh activation further bounds the pre-quantization values to $[-1,1]$, keeping the quantization error bounded and stabilizing training. In contrast, floating-point descriptors (e.g., SIFT, SuperPoint) undergo $L_2$ normalization. This preserves the original metric properties and ensures optimal performance when using Euclidean distance or cosine similarity for matching.

\noindent\textbf{Training Setup: } The model is optimized using the AdamW~\cite{b43} optimizer with a batch size of 16. We employ a cosine learning rate schedule, starting at $1\times10^{-3}$ after a 500-step linear warmup, and train it for 50 epochs. The entire training process for the final model required approximately 2 days on a single NVIDIA RTX 4090 GPU.


\begin{figure*}[t] 
    \centering
    \includegraphics[width=\textwidth]{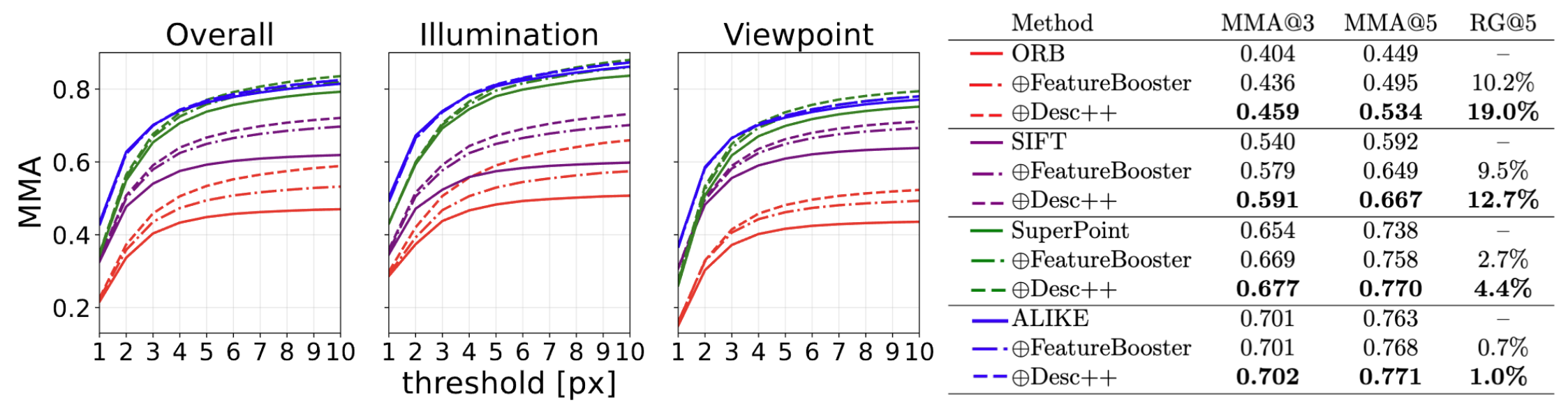} 
    \caption{MMA curves and quantitative results on the HPatches dataset. We compare Desc++ against the original descriptors and FeatureBooster. The table details MMA at 3px and 5px thresholds, with RG indicating the Relative Gain at 5px.} 
    \label{fig:Fig3} 
\end{figure*}


\begin{figure*}[t] 
    \centering
    \includegraphics[width=\textwidth]{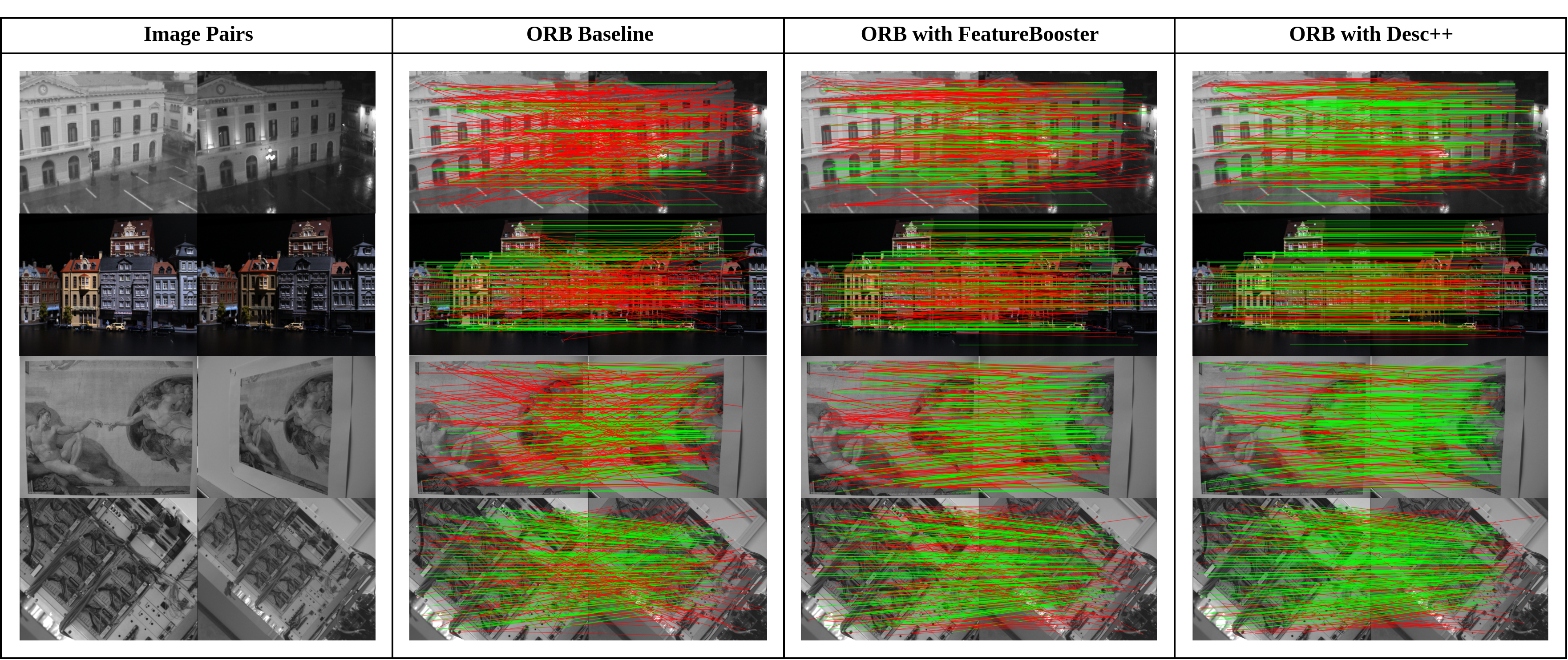} 
    \caption{Qualitative comparison of mutual nearest neighbor matching under illumination and viewpoint variations. Green and red lines denote correct (dist. $\le$ 5px) and incorrect correspondences, respectively.} 
    \label{fig:Fig4} 
\end{figure*}


\section{Experiment}
To validate the effectiveness of Desc++, we conduct experiments from both algorithmic and system perspectives. We first evaluate descriptor enhancement performance, followed by an analysis of its impact on visual data association. We then assess the resulting SLAM performance across multiple public benchmarks and compare the proposed method with representative learning-based front-ends in terms of both localization accuracy and deployment efficiency.

\begin{table*}[t]
\centering
\caption{Tracked MapPoints under different camera configurations. Camera-L, Camera-LS, and Camera-RS denote the left stereo camera, the left-side camera, and the right-side camera of the multi-camera rig, respectively; EuRoC and KITTI provide only the stereo pair, so results are reported on Camera-L. F: FeatureBooster~\cite{b7}; D: Desc++.}
\label{tab:tmp}

\scriptsize
\renewcommand{\arraystretch}{1.5}

\begin{tabular*}{\textwidth}{
    @{\extracolsep{\fill}}
    l
    cccccccccc
    @{}
}
\toprule
& \shortstack{EuRoC MH01}
& \shortstack{EuRoC V103}
& \shortstack{KITTI 01}
& \shortstack{KITTI 06}
& \multicolumn{3}{c}{Hilti-S1-H1}
& \multicolumn{3}{c}{Hilti-S3-H1} \\
\cmidrule(lr){6-8}
\cmidrule(lr){9-11}

Method
& Camera-L
& Camera-L
& Camera-L
& Camera-L
& Camera-L
& Camera-LS
& Camera-RS
& Camera-L
& Camera-LS
& Camera-RS \\
\midrule

Origin
& $270 \pm 86$
& $212 \pm 55$
& $342 \pm 63$
& $321 \pm 101$
& $282 \pm 148$
& $90 \pm 65$
& $79 \pm 59$
& $389 \pm 134$
& $109 \pm 73$
& $128 \pm 63$ \\

w/ F~\cite{b7}
& $279 \pm 85$
& $213 \pm 56$
& $345 \pm 63$
& $342 \pm 104$
& $246 \pm 125$
& $121 \pm 81$
& $103 \pm 71$
& $335 \pm 110$
& $163 \pm 97$
& $174 \pm 74$ \\

w/ D
& $281 \pm 82$
& $216 \pm 55$
& $353 \pm 59$
& $370 \pm 96$
& $274 \pm 134$
& $150 \pm 94$
& $127 \pm 86$
& $349 \pm 110$
& $189 \pm 107$
& $203 \pm 88$ \\

\bottomrule
\end{tabular*}
\end{table*}

\begin{figure*}[t] 
    \centering
    \includegraphics[width=\textwidth]{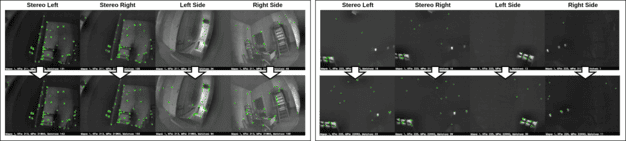} 
    \caption{Qualitative feature correspondences on the Hilti sequences. Desc++ preserves more valid feature-to-map associations under large cross-camera viewpoint changes.} 
    \label{fig:Fig5} 
\end{figure*}

\subsection{Experiment Setup}
This section introduces the datasets, evaluation metrics, and baseline methods used to comprehensively evaluate Desc++ at the descriptor, data association, and system levels.

\subsubsection{Dataset}
To evaluate whether improved descriptor quality translates into better visual data association and system-level SLAM performance, we conducted experiments on one descriptor benchmark and three representative SLAM benchmarks. Descriptor enhancement performance was first evaluated on the HPatches benchmark~\cite{b31}, which is a standard benchmark for assessing local descriptor robustness under significant viewpoint and illumination changes. Following the evaluation protocol of~\cite{b7,b35}, eight high-resolution sequences were excluded to ensure a fair comparison with previous descriptor enhancement methods. To evaluate the system-level impact of improved data association, experiments were further conducted on three widely used V-SLAM benchmarks. The EuRoC MAV dataset~\cite{b33} was used to evaluate performance in indoor environments. The KITTI Odometry dataset~\cite{b32} was adopted to assess large-scale outdoor localization. Finally, the Hilti SLAM Challenge 2023 dataset~\cite{b40} was included to evaluate robustness in challenging industrial environments with multi-camera configurations and large viewpoint changes.

\subsubsection{Evaluation Metrics}
We evaluate Desc++ from three complementary perspectives: descriptor quality, data association quality, and system-level SLAM performance. Descriptor enhancement performance is evaluated using the Mean Matching Accuracy (MMA)~\cite{b5} on the HPatches benchmark~\cite{b31}, which measures the percentage of correctly matched correspondences under different pixel error thresholds. A higher MMA indicates more discriminative and robust feature descriptors. To directly assess the impact on visual data association, we report the average number of Tracked MapPoints (TMP), defined as the number of local map points that are successfully associated with the current frame after descriptor matching and geometric verification. Higher TMPs indicate more reliable visual data association and provide additional geometric constraints for downstream SLAM optimization, which may improve camera tracking and pose estimation. System-level localization performance is evaluated using the Absolute Trajectory Error (ATE) Root Mean Square Error (RMSE) on all SLAM benchmarks with the EVO tool~\cite{b34}, where lower values indicate higher localization accuracy. Since modern V-SLAM systems exhibit stochastic behavior due to multi-thread scheduling, random initialization, and asynchronous optimization, each experiment is repeated five times. We report both the mean and standard deviation of the ATE RMSE to evaluate not only localization accuracy but also system robustness and run-to-run stability.

\subsubsection{Baselines}
We compare Desc++ with representative methods from three categories. First, we evaluate against the original feature-based V-SLAM systems, including ORB-SLAM2~\cite{b2}, ORB-SLAM3~\cite{b3}, RGB-L~\cite{b4}, and MAVIS-SLAM~\cite{b41}, which serve as the deployment baselines without descriptor enhancement. Second, FeatureBooster~\cite{b7}, the current state-of-the-art descriptor enhancement method, is adopted to evaluate improvements over existing descriptor refinement approaches under the same SLAM pipelines. Finally, Rover-SLAM~\cite{b42}, which employs SuperPoint~\cite{b8} and LightGlue~\cite{b13} to replace the original front-end, is included as a representative learning-based V-SLAM system for analyzing the trade-off between localization accuracy, computational efficiency, and deployment cost.

\subsection{Descriptor Matching Accuracy}
\label{sec:mma}
Figure~\ref{fig:Fig3} presents the quantitative descriptor matching performance on the HPatches benchmark~\cite{b31}, while Fig.~\ref{fig:Fig4} provides qualitative matching examples under challenging viewpoint and illumination changes. Desc++ consistently outperformed both the original descriptors and the previous descriptor enhancement method, FeatureBooster~\cite{b7}, across all evaluated descriptors. Notable improvements were observed for handcrafted features, where Desc++ achieved relative gains (RG@5) of 19.0\% and 12.7\% for ORB and SIFT, respectively, substantially exceeding those of FeatureBooster (10.2\% and 9.5\%). Desc++ also consistently improved learning-based descriptors, increasing SuperPoint's MMA@5 to 0.770 (+4.4\%) and ALIKE's to 0.771 (+1.0\%). As illustrated in Fig.~\ref{fig:Fig4}, the enhanced ORB produced denser and more reliable correspondences under severe viewpoint and illumination variations, qualitatively demonstrating the improved descriptor discriminability achieved by Desc++.

\begin{table*}[t]
\centering
\caption{ATE RMSE (cm) on the EuRoC~\cite{b33} dataset. The value below each result indicates the relative gain over ORB-SLAM2~\cite{b2} (stereo mode) and ORB-SLAM3~\cite{b3} (stereo-inertial mode). Lower values indicate better performance. OS2: ORB-SLAM2~\cite{b2}; OS3: ORB-SLAM3~\cite{b3}; F: FeatureBooster~\cite{b7}; D: Desc++; RS: Rover-SLAM~\cite{b42}. Bold and underlined values denote the best and second-best results, respectively. $\uparrow$/$\downarrow$: relative improvement/degradation with respect to the baseline. $\times$: tracking failure. Gains are computed from full-precision values and may differ slightly from ratios of the rounded values shown.}
\label{tab:euroc}
\fontsize{6.5}{7.8}\selectfont
\setlength{\tabcolsep}{1.2pt}
\renewcommand{\arraystretch}{1.15}

\begin{tabular*}{\textwidth}{
@{\extracolsep{\fill}}
lccccccccccc
@{}
}
\toprule
Method &
MH01 & MH02 & MH03 & MH04 & MH05 &
V101 & V102 & V103 & V201 & V202 & V203 \\
\midrule

OS2~\cite{b2} &
\gaincell{\plain{3.69\pm0.12}}{--} &
\gaincell{\plain{3.88\pm0.17}}{--} &
\gaincell{\second{4.04\pm0.08}}{--} &
\gaincell{\plain{10.11\pm2.52}}{--} &
\gaincell{\plain{5.18\pm0.39}}{--} &
\gaincell{\plain{8.72\pm0.08}}{--} &
\gaincell{\plain{6.48\pm0.09}}{--} &
\gaincell{\plain{7.90\pm0.69}}{--} &
\gaincell{\second{6.57\pm0.55}}{--} &
\gaincell{\second{5.94\pm0.50}}{--} &
\gaincell{\fail}{--} \\

w/ F~\cite{b7} &
\gaincell{\second{3.62\pm0.20}}{\posgain{1.92}} &
\gaincell{\second{3.87\pm0.23}}{\posgain{0.37}} &
\gaincell{\plain{4.24\pm0.66}}{\neggain{4.97}} &
\gaincell{\second{10.02\pm4.37}}{\posgain{0.91}} &
\gaincell{\best{4.32\pm0.27}}{\posgain{16.66}} &
\gaincell{\second{8.71\pm0.09}}{\posgain{0.06}} &
\gaincell{\best{6.32\pm0.06}}{\posgain{2.42}} &
\gaincell{\second{7.73\pm0.82}}{\posgain{2.19}} &
\gaincell{\plain{6.70\pm0.34}}{\neggain{2.07}} &
\gaincell{\plain{6.48\pm0.90}}{\neggain{9.09}} &
\gaincell{\fail}{--} \\

w/ D &
\gaincell{\best{3.47\pm0.17}}{\posgain{5.84}} &
\gaincell{\best{3.48\pm0.18}}{\posgain{10.40}} &
\gaincell{\best{3.86\pm0.28}}{\posgain{4.38}} &
\gaincell{\best{9.15\pm2.17}}{\posgain{9.46}} &
\gaincell{\second{4.73\pm0.43}}{\posgain{8.61}} &
\gaincell{\best{8.65\pm0.08}}{\posgain{0.78}} &
\gaincell{\second{6.44\pm0.62}}{\posgain{0.57}} &
\gaincell{\best{7.50\pm1.36}}{\posgain{5.03}} &
\gaincell{\best{6.02\pm0.44}}{\posgain{8.33}} &
\gaincell{\best{5.63\pm0.31}}{\posgain{5.14}} &
\gaincell{\fail}{--} \\

\midrule

OS3~\cite{b3} &
\gaincell{\plain{4.27\pm0.29}}{--} &
\gaincell{\plain{3.67\pm0.51}}{--} &
\gaincell{\plain{3.29\pm0.18}}{--} &
\gaincell{\second{4.58\pm0.12}}{--} &
\gaincell{\second{5.27\pm0.45}}{--} &
\gaincell{\second{3.73\pm0.14}}{--} &
\gaincell{\plain{1.79\pm0.06}}{--} &
\gaincell{\plain{2.55\pm0.14}}{--} &
\gaincell{\plain{3.56\pm0.35}}{--} &
\gaincell{\plain{1.46\pm0.36}}{--} &
\gaincell{\plain{4.17\pm0.88}}{--} \\

w/ F~\cite{b7} &
\gaincell{\plain{4.27\pm0.50}}{\posgain{0.14}} &
\gaincell{\best{2.81\pm0.58}}{\posgain{23.25}} &
\gaincell{\plain{3.28\pm0.17}}{\posgain{0.12}} &
\gaincell{\plain{4.60\pm0.42}}{\neggain{0.44}} &
\gaincell{\plain{5.56\pm0.71}}{\neggain{5.58}} &
\gaincell{\plain{3.75\pm0.05}}{\neggain{0.58}} &
\gaincell{\plain{1.83\pm0.19}}{\neggain{2.34}} &
\gaincell{\plain{2.54\pm0.16}}{\posgain{0.55}} &
\gaincell{\plain{3.65\pm0.26}}{\neggain{2.45}} &
\gaincell{\best{1.24\pm0.07}}{\posgain{15.00}} &
\gaincell{\plain{3.27\pm2.12}}{\posgain{21.66}} \\

w/ D &
\gaincell{\second{4.16\pm0.26}}{\posgain{2.57}} &
\gaincell{\second{2.96\pm0.24}}{\posgain{19.21}} &
\gaincell{\second{3.22\pm0.16}}{\posgain{1.89}} &
\gaincell{\best{4.37\pm0.14}}{\posgain{4.53}} &
\gaincell{\plain{5.39\pm0.64}}{\neggain{2.37}} &
\gaincell{\plain{3.77\pm0.06}}{\neggain{0.96}} &
\gaincell{\second{1.74\pm0.10}}{\posgain{2.55}} &
\gaincell{\second{2.33\pm0.10}}{\posgain{8.68}} &
\gaincell{\second{3.47\pm0.24}}{\posgain{2.54}} &
\gaincell{\second{1.31\pm0.07}}{\posgain{10.70}} &
\gaincell{\second{3.19\pm1.07}}{\posgain{23.64}} \\

\addlinespace[2pt]
\hdashline
\addlinespace[2pt]

RS~\cite{b42} &
\gaincell{\best{2.65\pm0.34}}{\posgain{37.90}} &
\gaincell{\plain{2.99\pm1.09}}{\posgain{18.52}} &
\gaincell{\best{3.00\pm0.71}}{\posgain{8.66}} &
\gaincell{\plain{4.91\pm0.24}}{\neggain{7.10}} &
\gaincell{\best{5.04\pm1.00}}{\posgain{4.32}} &
\gaincell{\best{3.44\pm0.12}}{\posgain{7.72}} &
\gaincell{\best{1.55\pm0.20}}{\posgain{13.40}} &
\gaincell{\best{2.07\pm0.25}}{\posgain{18.76}} &
\gaincell{\best{2.14\pm0.51}}{\posgain{39.89}} &
\gaincell{\plain{1.47\pm0.22}}{\neggain{0.24}} &
\gaincell{\best{3.01\pm1.14}}{\posgain{27.85}} \\

\bottomrule
\end{tabular*}
\end{table*}

\subsection{Data Association Analysis}
To investigate whether improved descriptor quality translates into more reliable visual data association, we analyze the number of Tracked MapPoints (TMPs) across representative operating scenarios. The selected sequences cover diverse environments across three public SLAM datasets, including a machine hall (EuRoC MH01), an indoor room (EuRoC V103), highway driving (KITTI-01), urban driving (KITTI-06), a construction site (Hilti S1-H1), and a warehouse environment (Hilti S3-H1). These scenarios span different sensing configurations and environmental characteristics, providing representative evidence of data association quality.

As summarized in Table~\ref{tab:tmp}, Desc++ generally increases the number of TMPs across the evaluated scenarios compared with both the original SLAM systems and  FeatureBooster~\cite{b7}. The improvement is most pronounced in the Hilti multi-camera sequences, where the side cameras are mounted at approximately $\pm 90^\circ$ relative to the primary camera and therefore observe the scene from near-orthogonal viewpoints, in contrast to the forward-facing stereo pair whose two views are nearly parallel. For example, on Hilti S1-H1, Desc++ increases the average TMPs from 90 to 150 on Camera-LS and from 79 to 127 on Camera-RS, with similar trends on S3-H1. Fig.~\ref{fig:Fig5} qualitatively illustrates this behavior, where the enhanced descriptors preserve more valid feature-to-map associations across cameras with substantially different viewpoints. Interestingly, the gains concentrate on the side cameras, while the TMPs on the primary camera (Camera-L) of the Hilti sequences remain at or slightly below the original system. A plausible explanation is that the primary camera observes the scene from viewpoints close to those from which the map points were created, so that the original descriptors already suffice for association, whereas the side cameras must associate features against map points created under markedly different viewing conditions. This within-sequence contrast is the most controlled comparison available, since the primary and side cameras share the same trajectory, scene, and SLAM system: the side-camera gains correspond to $+67\%$ and $+61\%$ on S1-H1 and $+73\%$ and $+59\%$ on S3-H1, whereas the primary camera changes by $-3\%$ and $-10\%$, respectively. At the trajectory level, Desc++ improves all nine Hilti sequences, and the benchmark-level improvement is statistically significant (Sec.~\ref{sec:significance}). We note that the present evaluation covers two discrete rig configurations rather than a continuum of viewpoint disparities, and we therefore report this as a comparison between the two configurations rather than as a relationship between disparity magnitude and gain.

These results suggest that improving descriptor discriminability facilitates more reliable feature-to-map associations, particularly in the side-camera configuration, allowing a larger number of map points to be successfully tracked in the current frame. Although TMPs is an intermediate indicator rather than a direct measure of localization accuracy, the observed improvements suggest that enhanced descriptors impose richer geometric constraints on subsequent SLAM optimization. This trend is further reflected in the trajectory accuracy reported in the following subsection.

\begin{table*}[t]
\centering
\caption{ATE RMSE (m) on the KITTI odometry dataset~\cite{b32}. The value below each result indicates the relative gain over RGB-L~\cite{b4} (stereo-LiDAR mode). Lower values indicate better performance. LS: RGB-L~\cite{b4}; F: FeatureBooster~\cite{b7}; D: Desc++. Bold and underlined values denote the best and second-best results, respectively. $\uparrow$/$\downarrow$: relative improvement/degradation with respect to the baseline. Gains are computed from full-precision values and may differ slightly from ratios of the rounded values shown.}
\label{tab:kitti}
\fontsize{6.5}{7.8}\selectfont
\setlength{\tabcolsep}{1.2pt}
\renewcommand{\arraystretch}{1.15}

\begin{tabular*}{\textwidth}{
@{\extracolsep{\fill}}
lccccccccccc
@{}
}
\toprule
Method &
00 & 01 & 02 & 03 & 04 & 05 &
06 & 07 & 08 & 09 & 10 \\
\midrule

LS~\cite{b4} &
\gaincell{\plain{5.47\pm0.25}}{--} &
\gaincell{\plain{110.44\pm36.64}}{--} &
\gaincell{\plain{10.86\pm1.67}}{--} &
\gaincell{\best{15.33\pm0.19}}{--} &
\gaincell{\plain{2.31\pm0.63}}{--} &
\gaincell{\plain{2.74\pm0.20}}{--} &
\gaincell{\second{4.23\pm0.26}}{--} &
\gaincell{\plain{0.84\pm0.04}}{--} &
\gaincell{\plain{16.86\pm0.95}}{--} &
\gaincell{\plain{6.49\pm0.61}}{--} &
\gaincell{\plain{7.54\pm0.57}}{--} \\

w/ F~\cite{b7} &
\gaincell{\best{5.44\pm0.16}}{\posgain{0.63}} &
\gaincell{\second{104.35\pm18.47}}{\posgain{5.52}} &
\gaincell{\second{8.94\pm0.96}}{\posgain{17.68}} &
\gaincell{\second{16.54\pm0.16}}{\neggain{7.93}} &
\gaincell{\second{1.69\pm0.34}}{\posgain{26.55}} &
\gaincell{\second{2.64\pm0.09}}{\posgain{3.51}} &
\gaincell{\plain{4.30\pm0.33}}{\neggain{1.61}} &
\gaincell{\second{0.81\pm0.06}}{\posgain{3.34}} &
\gaincell{\second{16.59\pm0.67}}{\posgain{1.58}} &
\gaincell{\best{5.54\pm0.20}}{\posgain{14.62}} &
\gaincell{\second{6.37\pm0.31}}{\posgain{15.54}} \\

w/ D &
\gaincell{\second{5.47\pm0.15}}{\posgain{0.16}} &
\gaincell{\best{99.85\pm13.33}}{\posgain{9.59}} &
\gaincell{\best{7.87\pm0.98}}{\posgain{27.55}} &
\gaincell{\plain{16.74\pm0.16}}{\neggain{9.21}} &
\gaincell{\best{1.45\pm0.12}}{\posgain{37.19}} &
\gaincell{\best{2.45\pm0.24}}{\posgain{10.52}} &
\gaincell{\best{3.54\pm0.48}}{\posgain{16.31}} &
\gaincell{\best{0.80\pm0.05}}{\posgain{5.52}} &
\gaincell{\best{16.44\pm0.96}}{\posgain{2.45}} &
\gaincell{\second{5.83\pm0.51}}{\posgain{10.11}} &
\gaincell{\best{5.71\pm0.53}}{\posgain{24.20}} \\

\bottomrule
\end{tabular*}
\end{table*}

\begin{table*}[t]
\centering
\caption{ATE RMSE (m) on the Hilti SLAM challenge~\cite{b40}. The value below each
result indicates the relative gain over MAVIS-SLAM~\cite{b41} (multi-cam inertial mode). Lower values indicate better performance. MS: MAVIS-SLAM~\cite{b41}; F: FeatureBooster~\cite{b7}; D: Desc++. Bold and underlined values denote the best and second-best results, respectively. $\uparrow$/$\downarrow$: relative improvement/degradation with respect to the baseline. Gains are computed from full-precision values and may differ slightly from ratios of the rounded values shown.}
\label{tab:hilti}
\fontsize{6.5}{7.8}\selectfont
\setlength{\tabcolsep}{1.2pt}
\renewcommand{\arraystretch}{1.15}

\begin{tabular*}{\textwidth}{
@{\extracolsep{\fill}}
lccccccccc
@{}
}
\toprule
Method &
S1-H1 & S1-H2 & S1-H3 & S1-H4 & S1-H5 &
S3-H1 & S3-H2 & S3-H3 & S3-H4 \\
\midrule

MS~\cite{b41} &
\gaincell{\plain{0.37\pm0.03}}{--} &
\gaincell{\plain{0.23\pm0.02}}{--} &
\gaincell{\second{0.34\pm0.02}}{--} &
\gaincell{\plain{0.23\pm0.04}}{--} &
\gaincell{\plain{0.41\pm0.02}}{--} &
\gaincell{\second{0.15\pm0.02}}{--} &
\gaincell{\plain{0.32\pm0.01}}{--} &
\gaincell{\plain{0.28\pm0.03}}{--} &
\gaincell{\second{0.23\pm0.03}}{--} \\

w/ F~\cite{b7} &
\gaincell{\second{0.33\pm0.06}}{\posgain{11.93}} &
\gaincell{\second{0.18\pm0.03}}{\posgain{20.97}} &
\gaincell{\plain{0.37\pm0.03}}{\neggain{10.39}} &
\gaincell{\second{0.22\pm0.03}}{\posgain{4.65}} &
\gaincell{\second{0.37\pm0.05}}{\posgain{10.36}} &
\gaincell{\plain{0.16\pm0.02}}{\neggain{5.81}} &
\gaincell{\second{0.31\pm0.01}}{\posgain{3.39}} &
\gaincell{\second{0.27\pm0.02}}{\posgain{4.64}} &
\gaincell{\plain{0.26\pm0.03}}{\neggain{13.58}} \\

w/ D &
\gaincell{\best{0.23\pm0.03}}{\posgain{37.31}} &
\gaincell{\best{0.16\pm0.03}}{\posgain{28.39}} &
\gaincell{\best{0.34\pm0.01}}{\posgain{0.87}} &
\gaincell{\best{0.20\pm0.01}}{\posgain{14.79}} &
\gaincell{\best{0.34\pm0.02}}{\posgain{18.24}} &
\gaincell{\best{0.13\pm0.02}}{\posgain{12.76}} &
\gaincell{\best{0.30\pm0.01}}{\posgain{5.91}} &
\gaincell{\best{0.26\pm0.04}}{\posgain{6.85}} &
\gaincell{\best{0.22\pm0.03}}{\posgain{2.36}} \\

\bottomrule
\end{tabular*}
\end{table*}

\subsection{SLAM Evaluation}
\subsubsection{EuRoC Dataset}
Table~\ref{tab:euroc} reports the results on the EuRoC~\cite{b33} dataset under both stereo (ORB-SLAM2~\cite{b2}) and stereo-inertial (ORB-SLAM3~\cite{b3}) configurations. In stereo mode, Desc++ reduced the ATE RMSE relative to ORB-SLAM2~\cite{b2} on all ten sequences that the systems completed, achieving the best result on eight of them. In particular, all methods, including the baseline and FeatureBooster~\cite{b7}, failed on V203 due to severe motion blur, which prevented the feature extractor from detecting meaningful patterns; we analyze this boundary case in Section~\ref{sec:discussion}. The purely visual configuration also highlights the stability benefit of Desc++: whereas FeatureBooster~\cite{b7} degraded the baseline on MH03, V201, and V202, Desc++ improved every completed sequence.

In stereo-inertial mode, Desc++ reduced the ATE RMSE relative to ORB-SLAM3~\cite{b3} on 9 of the 11 sequences, with relative improvements ranging from 1.9\% to 23.6\%; the largest gains were observed on MH02, V103, V202, and V203. Compared with FeatureBooster~\cite{b7}, Desc++ produced lower trajectory errors on 8 of the 11 sequences and exhibited more consistent behavior: it avoids the degradation of FeatureBooster on V201 ($+2.54\%$ versus $-2.45\%$), reduces the regression on MH05 ($-2.37\%$ versus $-5.58\%$), and shows substantially lower run-to-run variance on V203 ($\pm 1.07$ versus $\pm 2.12$). The only Desc++ regressions relative to the baseline occurred on MH05 and V101 ($-2.37\%$ and $-0.96\%$, respectively), both minor compared with the improvements on the remaining sequences, indicating that Desc++ improves localization accuracy overall. Nevertheless, the stereo-only results above reflect the improvement from descriptor quality more evidently.

\subsubsection{KITTI Dataset}
\label{sec:kitti}
As shown in Table~\ref{tab:kitti}, Desc++ achieved the lowest ATE RMSE on 8 of the 11 KITTI~\cite{b32} sequences and outperformed FeatureBooster~\cite{b7} on 8 sequences. The largest relative gains over RGB-L~\cite{b4} were obtained on Sequences 04 (37.19\%), 02 (27.55\%), and 10 (24.20\%), and the improvements on Sequences 02, 04, 06, and 10 are statistically significant (Sec.~\ref{sec:significance}). Sequences 04 and 10 contain no loop closures, so their accumulated drift is not corrected by the back-end and depends largely on the quality of local feature associations; the gains there suggest that improved descriptor discriminability provides more reliable geometric constraints against drift. Sequence 01 deserves particular attention: this highway sequence combines high vehicle speed with sparse, distant features and is known to challenge feature-based systems, as reflected in the large baseline error and run-to-run variance ($110.44 \pm 36.64$\,m). While the absolute error remains large, Desc++ not only reduced the mean drift but also substantially reduced run-to-run variance ($\pm 13.33$\,m versus $\pm 36.64$\,m), indicating more stable data association under these adverse conditions. The main regression occurred on Sequence 03, where Desc++ fell behind the RGB-L~\cite{b4} baseline by 9.21\%; this is the only statistically significant regression among all sequence-level comparisons in this work. FeatureBooster~\cite{b7} shows a comparable degradation on the same sequence (7.93\%), suggesting a property of this sequence rather than a failure mode specific to our method. In absolute terms, the gap is 1.41 m, and the run-to-run variance of Desc++ on this sequence is low ($\pm0.16$\,m), indicating a systematic offset rather than unstable behavior. Nevertheless, the overall trend across the benchmark remained positive, demonstrating that the proposed enhancement generalizes to large-scale outdoor environments.

\subsubsection{Hilti SLAM Challenge 2023}
As shown in Table~\ref{tab:hilti}, Desc++ achieved the lowest ATE RMSE on all nine Hilti Challenge~\cite{b40} sequences, outperforming both the original MAVIS-SLAM~\cite{b41} and the FeatureBooster-enhanced version. Relative improvements over MAVIS-SLAM~\cite{b41} ranged from 0.9\% to 37.3\%, with particularly large gains on S1-H1, S1-H2, and S1-H5. While the margins on some sequences are within the run-to-run variance, Desc++ is the only method that improves upon the baseline on every sequence, whereas FeatureBooster~\cite{b7} degrades performance on S1-H3, S3-H1, and S3-H4. Compared with the EuRoC~\cite{b33} and KITTI~\cite{b32} benchmarks, the Hilti Challenge~\cite{b40} poses a more demanding multi-camera SLAM scenario, where feature correspondences must be established across cameras with substantially different viewpoints and smaller overlapping fields of view. Under these conditions, the quality of local descriptors plays a more critical role in maintaining reliable data association, consistent with the TMPs analysis in Table~\ref{tab:tmp}. The improved localization accuracy across all sequences, therefore, suggests that enhancing descriptor discriminability is particularly beneficial for multi-camera visual-inertial SLAM, where robust feature matching across heterogeneous viewpoints is essential for accurate state estimation.

\subsubsection{Statistical Significance Analysis}
\label{sec:significance}
Since multi-threaded SLAM systems exhibit run-to-run variability, we assess whether the observed improvements exceed this variability at two levels. First, for each sequence, we compare the five runs of Desc++ against the five runs of the corresponding baseline using a two-sided Welch's t-test. Second, for each system--dataset combination, we perform a paired Wilcoxon signed-rank test over the per-sequence mean ATE RMSE values, which assesses whether the improvement is consistent across an entire benchmark. With five runs per method, these per-sequence tests have limited statistical power; we therefore treat them as descriptive and do not correct them for multiple comparisons, basing our inferential claims on the benchmark-level Wilcoxon tests.

As summarized in Table~\ref{tab:significance}, the Wilcoxon tests confirm that the overall improvement of Desc++ is statistically significant in all four configurations ($p \le 0.024$). At the per-sequence level, eleven comparisons show individually significant gains ($p<0.05$), including EuRoC (MH02, MH04, V103), KITTI (02, 04, 06, 10), and the Hilti Challenge (S1-H1, S1-H2, S1-H5). Improvements below approximately $3\%$ generally fall within run-to-run variability, and we therefore do not claim per-sequence significance for them; however, the corresponding regressions are equally non-significant. A plausible explanation is that, on these sequences, the baseline already establishes sufficient associations for accurate estimation, so the residual error is dominated by factors other than descriptor quality, and further improvement in matching yields little measurable benefit; we have not isolated these factors directly. Across all 41 sequence-level comparisons, the only statistically significant regression is KITTI Sequence~03 ($p<0.001$), discussed in Section~\ref{sec:kitti}, on which FeatureBooster exhibits a comparable degradation. Together, these results indicate that the improvement is consistent across sequences rather than being confined to a small number of favorable cases.

\begin{table}[t]
\centering
\caption{Statistical significance of Desc++ over each baseline. Per-sequence: two-sided Welch's t-test over five runs ($p<0.05$). Overall: paired Wilcoxon signed-rank test across sequences.}
\label{tab:significance}
\fontsize{6.5}{7.8}\selectfont
\setlength{\tabcolsep}{1.2pt}
\renewcommand{\arraystretch}{1.15}
\begin{tabular*}{\columnwidth}{@{\extracolsep{\fill}}lccc@{}}
\toprule
Configuration & Improved & Significant seq. & Wilcoxon $p$ \\
\midrule
ORB2 (EuRoC)  & 10/10 & MH02                & 0.002 \\
ORB3 (EuRoC)  &  9/11 & MH02, MH04, V103    & 0.019 \\
RGB-L (KITTI) & 10/11 & 02, 04, 06, 10      & 0.024 \\
MAVIS (Hilti) &  9/9  & S1-H1, S1-H2, S1-H5 & 0.004 \\
\bottomrule
\end{tabular*}
\end{table}

\subsection{Comparison with learning-based front-ends}
To further position the proposed approach with respect to recent learning-based SLAM systems, we compare Desc++ with Rover-SLAM~\cite{b42}, a representative learning-based front-end built upon SuperPoint~\cite{b8} and LightGlue~\cite{b13}, and LF2SLAM~\cite{b38}, which replaces the ORB detector with a pose-estimation-tailored SuperPoint network while retaining the original binary descriptors. Each is tested on the same desktop platform (Intel i9-12900K, 128\,GB RAM, NVIDIA RTX 4090). Unlike the previous experiments, this comparison evaluates the trade-off between localization accuracy and computational cost rather than directly comparing individual descriptors. The localization accuracy is evaluated on EuRoC using the stereo-inertial configuration provided by the publicly available Rover-SLAM implementation, and the runtime cost is measured on the MH01 sequence. As the official LF2SLAM implementation targets the monocular configuration of ORB-SLAM3, it is included in the computational comparison only.

As shown in Table~\ref{tab:euroc}, Rover-SLAM achieved the lowest ATE RMSE on several sequences, demonstrating the strong localization capability of modern learning-based front-ends. Nevertheless, Desc++ remained competitive, achieving comparable accuracy on multiple sequences while improving upon the original ORB-SLAM3~\cite{b3}. Rover-SLAM's accuracy advantage, however, comes at a substantially higher computational cost. As summarized in Table~\ref{tab:runtime_cost}, Desc++ adds only 5.3\,ms to the front-end latency of ORB-SLAM3 and maintains real-time performance at 30\,FPS, whereas Rover-SLAM requires 71.9\,ms per frame and operates at 14\,FPS even on high-end hardware. LF2SLAM occupies an intermediate position: its front-end cost (33.3\,ms, 836\,MiB) is of the same order as Desc++, but this cost is incurred to replace the ORB detector, whereas Desc++ preserves the detector, the descriptor format, and the matching interface, enabling the same enhancement module to be integrated into four heterogeneous SLAM systems while preserving their existing front-end pipelines.

The memory footprint of Rover-SLAM is substantially larger than that of Desc++, at 6130\,MiB versus 544\,MiB. Although 6130\,MiB does not exceed the nominal 8\,GB memory capacity of the Jetson Orin NX, it leaves limited memory headroom because the available memory must be shared between the GPU and the rest of the SLAM system. In practice, Rover-SLAM could not be deployed on this platform. In contrast, the Desc++-enhanced system runs end-to-end at 20\,FPS on the Jetson Orin NX, confirming that the enhancement remains viable under embedded constraints. Regarding frame-rate headroom, the measured latency increases approximately linearly with the number of keypoints rather than quadratically, providing a predictable computational trade-off as the frame budget changes. On the desktop platform, the enhancement adds 5.3\,ms per frame, leaving substantial headroom within a 33\,ms budget at 30\,Hz. On the Jetson Orin NX, the enhanced system runs end-to-end at 20\,FPS under the default configuration; where a higher frame rate is required, the measured scaling in Table~\ref{tab:hpatches_latency} provides a direct trade-off, as reducing the number of enhanced keypoints to 500 lowers the module latency to approximately 10\,ms.

\begin{table}[t]
\centering
\caption{End-to-end computational cost in the integrated SLAM system. Latency is averaged per frame over EuRoC MH01. Bold denotes components that replace or augment the original ORB front-end, and a dash denotes that no GPU memory is used. LF2SLAM~\cite{b38} runs in monocular mode (its official configuration) and processes one image per frame, whereas the other rows process a stereo pair; its values are therefore not directly comparable and are included for reference on computational cost only. D: Desc++; SP: SuperPoint~\cite{b8}; SP(B): the SLAM-specific SuperPoint keypoint extractor of LF2SLAM~\cite{b38}, which retains the original binary descriptors; LG: LightGlue~\cite{b13}; GV: geometric verification.}
\label{tab:runtime_cost}
\scriptsize
\setlength{\tabcolsep}{3pt}
\renewcommand{\arraystretch}{1.25}
\begin{tabular*}{\columnwidth}{
l@{\extracolsep{\fill}}lccc
}
\toprule
Method & Front-end & Latency (ms) $\downarrow$ & Mem. (MiB) $\downarrow$ & FPS $\uparrow$ \\
\midrule
\multicolumn{5}{l}{Desktop (i9-12900K, RTX 4090)} \\
\midrule
OS3~\cite{b3}   & ORB + GV                   & $27.2 \pm 1.8$  & --   & 37 \\
OS3 w/ D        & ORB + \textbf{D} + GV      & $32.5 \pm 2.5$  & 544  & 30 \\
LF2~\cite{b38}  & \textbf{SP(B)} + GV        & $33.3 \pm 4.2$  & 836  & 30 \\
RS~\cite{b42}   & \textbf{SP} + \textbf{LG}  & $71.9 \pm 24.5$ & 6130 & 14 \\
\midrule
\multicolumn{5}{l}{Embedded (Jetson Orin NX, 8\,GB)} \\
\midrule
OS3~\cite{b3}   & ORB + GV                   & $37.6 \pm 6.1$  & --   & 27 \\
OS3 w/ D        & ORB + \textbf{D} + GV      & $50.8 \pm 8.4$  & 582  & 20 \\
RS~\cite{b42}   & \textbf{SP} + \textbf{LG}  & \multicolumn{3}{c}{Not deployable (insufficient memory)} \\
\bottomrule
\end{tabular*}
\end{table}

\subsection{Analysis of the Binary Descriptor}
\subsubsection{Hamming distance distributions}
\label{sec:hamming}

Matching accuracy alone does not show how the proposed descriptor reshapes the binary embedding space. Since ORB descriptors are confined to the vertices of the Hamming cube $\{0,1\}^{256}$, any improvement must manifest as a change in the distance statistics of the retrieved correspondences. For each keypoint, we compute its nearest neighbor (NN) under the Hamming metric on the full HPatches benchmark ($108$ sequences, $540$ pairs) and record the distance $d_1$. A retrieved correspondence is labeled correct if its reprojection error under the ground-truth homography is at most $3$\, px and incorrect if it exceeds $10$\,px; the intermediate band is discarded to avoid contaminating either mode, and keypoints projected outside the target image are excluded. Note that the incorrect mode does not follow the $\mathrm{Bin}(256,0.5)$ law of independent binary codes, since $d_1$ is a minimum over all candidates; both modes are extreme-value statistics characterizing the operating regime of the matcher.

\begin{figure}[t] 
    \centering
    \includegraphics[width=0.48\textwidth]{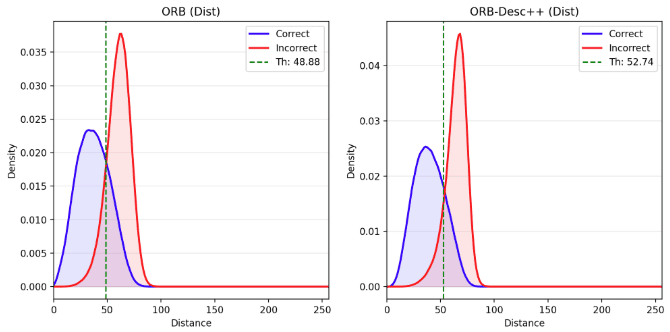} 
    \caption{Nearest-neighbor Hamming distance distributions on HPatches for (left) ORB and (right) ORB-Desc\texttt{++}. Blue and red denote correct and incorrect correspondences; the dashed line marks the density intersection $\tau^{*}$ of Table~\ref{tab:hamming}.} 
    \label{fig:hamming} 
\end{figure}

Three effects are apparent in Fig.~\ref{fig:hamming} and Table~\ref{tab:hamming}. First, the incorrect mode shifts toward larger distances and becomes sharper, so false candidates are rejected with a more consistent margin. Second, the discriminability $d'=(\mu^{-}\!-\!\mu^{+})/\sqrt{(\sigma^{+2}\!+\!\sigma^{-2})/2}$ improves by $14.0\%$, decomposing into a $5.8\%$ gain in inter-mode separation and a $7.2\%$ reduction in pooled dispersion; the gain is thus not a mere rescaling of the distance axis. Third, the correct mode also shifts slightly. This is a composition effect rather than a degradation: the correct population grows by $53.0\%$ while the incorrect one shrinks by $27.2\%$, and the newly recovered
correspondences are precisely the difficult ones the baseline failed to retrieve, whose distances are inherently larger. Separability nonetheless improves, with the overlap $\int\!\min(p^{+}\!(d),p^{-}\!(d))\,\mathrm{d}d$ falling by $16.5\%$ and the verification error $1\!-\!\mathrm{AUC}$ by $24.7\%$.

Consequently, the density intersection $\tau^{*}$ moves from $48.88$ to $52.74$ bits, so a matcher can admit correspondences over a $7.9\%$ wider distance range at a comparable error rate, consistent with the recall gain in Section~\ref{sec:pr}. Since the two populations are unequally sized and are defined by each method's own retrieved neighbors, $\tau^{*}$ is reported as a descriptive statistic rather than a Bayes-optimal boundary.

\subsubsection{Quantization error and code efficiency under STE}
\label{sec:quant}

Quantization in our design affects only the backward pass: binarization occurs before the loss is evaluated (Sec.~\ref{sec:train}), so the training objective is computed on the exact codes used at inference, and the optimized distance is the true Hamming distance, leaving no soft-to-hard transfer gap. What the estimator approximates is the gradient alone, replacing the a.e.-zero derivative of the sign function with that of the identity surrogate $x\mapsto x$, whose tightness is governed by how far the pre-binarization outputs lie from the vertices of the cube. Over all $1.92\times10^{6}$ HPatches descriptors we measure a quantization residual $\varepsilon=\mathbb{E}[\,|b-x|\,]=\mathbb{E}[\,1-|x|\,]=0.119$; the mean understates the saturation, as the distribution of $|x|$ is strongly left-skewed with a median of $0.998$ and only $2.4\%$ of bits falling within $0.1$ of the decision boundary. The surrogate is therefore evaluated where it best approximates the sign function for most bits, and the quantization error is concentrated in a small minority.

What ultimately matters, however, is whether the resulting hard codes carry usable information. ORB already addresses part of this: rBRIEF selects its $256$ tests by greedily maximizing variance and minimizing pairwise correlation, which our measurements confirm, as no bit is degenerate and the mean per-bit entropy reaches $0.989$. The selection is nevertheless performed once, offline, and applied to every patch, so content-dependent redundancy remains: the mean absolute inter-bit correlation is $0.105$ and the effective dimension is only $d_{\text{eff}}=39.0$, i.e.\ the nominal $256$ bits carry the variance of roughly $39$ independent ones. Because our descriptor re-encodes each patch individually, training through the estimator reduces the mean correlation by $35.2\%$ and nearly doubles $d_{\text{eff}}$ to $71.3$ (Table~\ref{tab:bitstats}), showing that the gradient approximation does not prevent the network from learning a substantially more efficient binary representation.

\begin{table}[t]
\centering
\caption{Nearest-neighbor Hamming distance statistics on HPatches ($540$ image pairs). $N^{\pm}$ are the numbers of correct/incorrect retrieved correspondences and $\tau^{*}$ the density intersection.}
\label{tab:hamming}
\footnotesize
\setlength{\tabcolsep}{2pt}
\begin{tabularx}{\columnwidth}{@{}l >{\centering\arraybackslash}X >{\centering\arraybackslash}X >{\raggedleft\arraybackslash}X@{}}
\toprule
 & ORB & Desc\texttt{++} & $\Delta$ \\
\midrule
$N^{+}$                    & $328.7$k & $\mathbf{502.8}$k & $+53.0\%$ \\
$N^{-}$                    & $1149.3$k & $\mathbf{836.9}$k & $-27.2\%$ \\
\midrule
$\mu^{+}\!\pm\!\sigma^{+}$ & $37.3\!\pm\!15.2$ & $39.9\!\pm\!14.2$ & -- \\
$\mu^{-}\!\pm\!\sigma^{-}$ & $60.2\!\pm\!11.0$ & $64.2\!\pm\!10.0$ & -- \\
\midrule
$d'\uparrow$               & $1.736$ & $\mathbf{1.979}$ & $+14.0\%$ \\
Overlap$\downarrow$        & $0.387$ & $\mathbf{0.323}$ & $-16.5\%$ \\
$1\!-\!$AUC$\downarrow$    & $0.116$ & $\mathbf{0.087}$ & $-24.7\%$ \\
$\tau^{*}$                 & $48.88$ & $52.74$ & $+7.9\%$ \\
\bottomrule
\end{tabularx}
\end{table}

\begin{table}[t]
\centering
\caption{Per-bit statistics of the $256$-bit codes over all HPatches descriptors ($1.92\times10^{6}$). $d_{\text{eff}}=(\sum_k\lambda_k)^2/ \sum_k\lambda_k^2$ is the participation ratio of the inter-bit correlation spectrum; an ideal code attains $d_{\text{eff}}=256$.}
\label{tab:bitstats}
\footnotesize
\begin{tabular*}{0.5\textwidth}{@{\extracolsep{\fill}}lccr@{}}
\toprule
 & ORB & Desc\texttt{++} enhanced ORB & $\Delta$ \\
\midrule
Mean $|\rho_{ij}|\downarrow$      & $0.105$ & $\mathbf{0.068}$ & $-35.2\%$ \\
$d_{\text{eff}}\uparrow$          & $39.0$  & $\mathbf{71.3}$  & $+82.6\%$ \\
\bottomrule
\end{tabular*}
\end{table}

\begin{figure}[t] 
    \centering
    \includegraphics[width=0.48\textwidth]{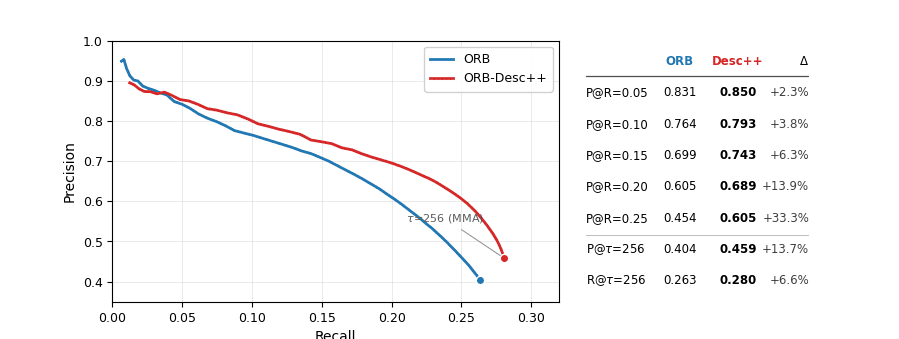} 
    \caption{Precision--recall curves on HPatches, obtained by sweeping the Hamming acceptance threshold $\tau$. Dotted lines mark the recall levels quoted in the text; the large markers indicate the $\tau=256$ operating point reported as matching accuracy.} 
    \label{fig:pr} 
\end{figure}

\begin{figure}[!t]
\centerline{\includegraphics[width=\columnwidth]{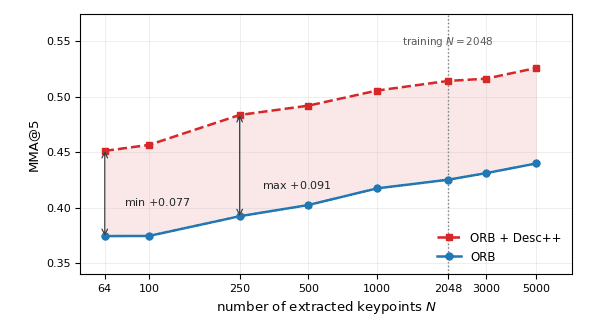}}
\caption{Matching accuracy as a function of the number of extracted ORB keypoints, on the 64 HPatches sequences unsaturated at $N=5000$. The shaded band is the gain from enhancement, which stays within $[+0.077, +0.091]$ over a 78-fold change in keypoint count and does not peak at the training configuration.}
\label{fig:kpt_sensitivity}
\end{figure}

\subsubsection{Precision--recall behavior}
\label{sec:pr}

The matching accuracy reported in Sec.~\ref{sec:mma} characterizes a single operating point: every mutual nearest-neighbor match is accepted, and the reported quantity is the precision of that fixed set. To expose the full trade-off, we keep the reprojection threshold at $3$\,px and instead sweep the Hamming acceptance threshold $\tau$, defining $\mathrm{P}(\tau)=\#\{\text{correct},\,d\!\leq\!\tau\}/\#\{d\!\leq\!\tau\}$ and $\mathrm{R}(\tau)=\#\{\text{correct},\,d\!\leq\!\tau\}/N_{\mathrm{GT}}$, where $N_{\mathrm{GT}}$ counts reference keypoints whose projection falls within $3$\, px of some target keypoint. Both statistics are averaged over the $540$ image pairs, consistent with the protocol of Sec.~\ref{sec:mma}; precision is averaged only over pairs retaining at least one match at the current threshold, since it is otherwise undefined. As the same detector is used throughout, $N_{\mathrm{GT}}$ is identical for both methods, and at $\tau=256$ the curve recovers exactly the operating point reported earlier ($0.404$ and $0.459$).
Figure~\ref{fig:pr} shows that the enhanced descriptor dominates the baseline over the whole range, with a margin that widens systematically towards higher recall. At $R=0.10$ the advantage is modest (ORB 0.764 vs. ORB-Desc++ 0.793), since at conservative thresholds both descriptors retain only their most confident matches; the gap then grows to $+6.3\%$ at $R=0.15$, $+13.9\%$ at $R=0.20$ and $+33.3\%$ at $R=0.25$. The baseline thus loses precision far more rapidly once the threshold is relaxed, whereas ORB-Desc\texttt{++} sustains a usable precision over a substantially wider recall range, which is precisely the regime in which practical matching pipelines operate. The same effect is visible in the composition of the match set at the fixed operating point, where ORB-Desc\texttt{++} returns $4.9\%$ fewer mutual nearest-neighbor matches ($511{,}590$ vs.\ $538{,}210$) yet recovers more correct correspondences (recall $0.263\!\rightarrow\!0.280$), confirming that the gain comes from better ranking rather than from admitting a larger candidate set. This wider margin is the matching-level manifestation of the distributional change reported in Sec.~\ref{sec:hamming}: the reduced overlap between the two Hamming modes allows a looser threshold without a proportionate loss in precision.

\begin{table*}[ht]
\centering
\caption{Module-level comparison of context aggregation backbones under a comparable parameter budget and identical hidden dimension. The number of layers is adjusted per backbone to meet this budget and is reported for completeness. Descriptor enhancement accuracy and inference latency are measured on RTX 4090 and Jetson Orin NX.}
\label{tab:hpatches_latency}
\vspace{1mm}

\renewcommand{\arraystretch}{1.5}
\fontsize{7}{7}\selectfont
\setlength{\tabcolsep}{8pt}{
\begin{tabular*}{\textwidth}{
@{\extracolsep{\fill}}
lccccc
@{}
}

\toprule

\textbf{Descriptor Enhancer} &
\makecell[c]{\textbf{MMA $\uparrow$}\\\textbf{@3 / @5}} &
\makecell[c]{\textbf{\#Params}$\downarrow$} &
\makecell[c]{\textbf{\#Layers}$\downarrow$} &
\makecell[c]{\textbf{RTX 4090  }\textbf{Runtime (ms) $\downarrow$}\\\textbf{\#500 / \#1k / \#2k / \#4k / \#8k}} &
\makecell[c]{\textbf{Jetson Orin NX  }\textbf{Runtime (ms) $\downarrow$}\\\textbf{\#500 / \#1k / \#2k / \#4k / \#8k}} \\

\midrule

AFT-Simple (FeatureBooster~\cite{b7})
&
0.436 / 0.495
&
2.61M
&
4
&
\textcolor{blue}{0.6} /
\textcolor{blue}{0.7} /
\textcolor{blue}{1.0} /
\textcolor{blue}{1.5} /
\textcolor{blue}{2.7}
&
\textcolor{blue}{7.0} /
\textcolor{blue}{9.7} /
\textcolor{blue}{18.9} /
\textcolor{red}{35.4} /
\textcolor{red}{67.7}
\\

Vanilla Transformer~\cite{b15}
&
0.437 / 0.500
&
2.61M
&
4
&
\textcolor{blue}{0.9} /
\textcolor{blue}{1.4} /
\textcolor{blue}{4.0} /
\textcolor{blue}{13.9} /
\textcolor{red}{51.9}
&
\textcolor{blue}{11.2} /
\textcolor{red}{35.6} /
\textcolor{red}{93.3} /
\textcolor{red}{340.0} /
\textcolor{red}{OOM}
\\

Mamba~\cite{b17}
&
0.443 / 0.510
&
2.60M
&
3
&
\textcolor{blue}{1.2} /
\textcolor{blue}{1.2} /
\textcolor{blue}{1.2} /
\textcolor{blue}{1.5} /
\textcolor{blue}{2.9}
&
\textcolor{blue}{9.0} /
\textcolor{blue}{11.8} /
\textcolor{blue}{21.8} /
\textcolor{red}{41.6} /
\textcolor{red}{80.6}
\\

\textbf{Mamba-AFTs}
&
\textbf{0.459 / 0.534}
&
\textbf{2.44M}
&
2
&
\textcolor{blue}{1.1} /
\textcolor{blue}{1.1} /
\textcolor{blue}{1.1} /
\textcolor{blue}{1.1} /
\textcolor{blue}{2.5}
&
\textcolor{blue}{10.1} /
\textcolor{blue}{11.3} /
\textcolor{blue}{21.0} /
\textcolor{red}{40.4} /
\textcolor{red}{77.5}
\\

\bottomrule

\end{tabular*}
}

\vspace{1mm}

\begin{minipage}{\textwidth}
\footnotesize
\textit{Note:}
\textcolor{blue}{Blue text} indicates real-time performance ($>$30 FPS),
\textcolor{red}{Red text} indicates sub-real-time performance ($<$30 FPS),
and \textbf{OOM} denotes out-of-memory.
\end{minipage}

\end{table*}

\subsubsection{Sensitivity to Keypoint Count}
Desc++ is trained up to $N=2048$ keypoints per image, but the number of features a detector returns varies considerably across frames in deployment. To assess whether the enhancement remains effective away from this configuration, we re-extract ORB features on HPatches with ORB-SLAM's octree-based extractor at each target count so the keypoint set matches what the front end would receive at that setting, then apply Desc++ without retraining. Fig.~\ref{fig:kpt_sensitivity} reports MMA@5 over the 64 sequences in which no image saturates the detector at $N=5000$, so that every reported count is attained exactly. The improvement is preserved throughout. Over a 78-fold variation in keypoint count, the gain in MMA@5 remains between $+0.077$ and $+0.091$, with no maximum at the training configuration; the number of correct matches is likewise higher at every setting (883.5 versus 796.0 per image pair at $N=5000$). The gain at the sparse end ($+0.077$ at $N=64$), the regime least represented during training, is comparable to that at $N=2048$ ($+0.089$). The full 108-sequence set, which includes images that saturate the detector at the larger settings, yields the same behavior.

This robustness follows from the module's design. Desc++ applies a single feed-forward mapping per frame and carries no state across frames, so a change in keypoint count between consecutive frames introduces no failure mode beyond the per-frame behavior characterized here. The range covered is representative of deployment: the ORB-SLAM3 configurations used on these benchmarks extract between 1000 and 2000 features per frame, well inside the stable region.

\subsection{Ablation Study}
\subsubsection{Context aggregation backbone}
We first compare the Mamba-AFTs block against alternative context aggregation architectures under a similar parameter budget ($\approx$2.6M). As shown in Table~\ref{tab:hpatches_latency}, Mamba-AFTs achieved higher ORB matching accuracy than AFT-Simple (FeatureBooster)~\cite{b7}, the Vanilla Transformer~\cite{b15}, and pure Mamba~\cite{b17}, while using slightly fewer parameters. Since the competing backbones have slightly more parameters than Mamba-AFTs, matching them exactly to 2.44M would only reduce their capacity; therefore, the performance advantage cannot be attributed to additional parameter capacity. In particular, it outperformed not only FeatureBooster's AFT-only design but also the pure Mamba variant, indicating that the two branches capture complementary context: neither order-agnostic global attention nor sequential state-space modeling alone matches their parallel combination. 

\subsubsection{Layer-wise interaction of the two branches}
\label{sec:gate_analysis}

To further understand the behavior and interactions of the two branches, we analyze the fusion weights over all 1.9M keypoints in the HPatches sequences and evaluate the contribution of each branch by overriding the gate at inference time (Table~\ref{tab:gate_ablation}). The two Context Aggregation layers behave differently. In the first layer, the fusion weight assigned to the AFT branch is nearly zero ($w_{\text{AFT}} < 1.6\times10^{-7}$), indicating that the layer predominantly relies on the order-sensitive Mamba branch. In contrast, the second layer adopts a stable mixture of the two branches, with $w_{\text{AFT}} = 0.584 \pm 0.063$.

The ablation results further demonstrate that this layer-wise specialization contributes to the final representation. Forcing the first layer to use uniform fusion reduces MMA@5 by 0.030. In the second layer, restricting the computation to either a single branch also degrades performance, with decreases of 0.022 for Mamba-only and 0.040 for AFT-only. These results indicate that the two branches contribute complementarily to the final representation, with the first layer relying primarily on the Mamba branch, while the second layer benefits from combining the Mamba and AFT branches. Although the second layer employs a learned fusion weight, its value remains relatively stable across inputs. Replacing the learned gate with a fixed uniform assignment produces no measurable change in MMA@5. Therefore, the observed performance benefit primarily arises from combining the two aggregation mechanisms rather than from fine-grained, input-dependent weight modulation.

\subsubsection{Runtime scalability}
In terms of runtime, Mamba-AFTs exhibited favorable linear scalability across hardware platforms. On an RTX 4090, it processed 8,000 keypoints in 2.5\,ms, whereas the Vanilla Transformer incurred substantially higher latency (51.9\,ms) due to its quadratic attention complexity. Similar trends were observed on the Jetson Orin NX, where Mamba-AFTs maintained real-time performance ($>$30\,FPS) under typical SLAM workloads (1k--2k keypoints), while the Transformer-based model encountered out-of-memory (OOM) failures. 

\subsubsection{Component-wise ablation}
The component-wise ablation in Table~\ref{tab:ablation_hpatches} examines the Local Fusion and Context Aggregation stages. Introducing the geometry encoder ($\mathrm{MLP}_{\text{geo}}$) improved matching performance by incorporating geometric cues, and replacing it with Learnable Fourier Features ($\mathrm{LFF}\text{-}\mathrm{MLP}_{\text{geo}}$) provided additional gains. The Context Aggregation stage contributed the largest single improvement, raising MMA@3 and MMA@5 to 0.459 and 0.534, respectively. These results indicate that each component contributes positively to the overall enhancement quality. 

\subsubsection{Serialization}
Table~\ref{tab:serialization_ablation} compares four strategies for ordering the keypoint set before it enters the Mamba branch. The comparison separates orderings that preserve two-dimensional locality from those that do not. Row-major scanning imposes a deterministic order but constrains adjacency along one axis; with sparse keypoints, its horizontal tie-break is rarely exercised, so the resulting sequence retains little two-dimensional structure while discarding the keypoint-response ordering present in the detector's native output. Row-major scanning therefore falls below the unordered baseline on both thresholds. Both locality-preserving curves, Hilbert and Z-order, recover or exceed that baseline, with Z-order performing best at each threshold. The gain over the unordered baseline is modest because it is a net figure: serialization must first discard the detector's native ordering to impose a spatial one. The parallel AFT branch is permutation-invariant and further compresses the aggregate effect, which is what makes the row-major result informative, since an unsuitable ordering degrades performance by a larger margin than a suitable ordering improves it.

\begin{table}[t]
\centering
\caption{Inference-time gate override on HPatches. Starting from the deployed model, we replace the softmax gate of a given Context Aggregation layer with a fixed assignment. No retraining is involved.}
\label{tab:gate_ablation}
\vspace{1mm}

\renewcommand{\arraystretch}{1.4}
\fontsize{7}{7}\selectfont
\setlength{\tabcolsep}{6pt}{
\begin{tabular*}{\columnwidth}{@{\extracolsep{\fill}}lccc@{}}
\toprule
\textbf{Gate configuration} &
\textbf{MMA@3 $\uparrow$} &
\textbf{MMA@5 $\uparrow$} &
\textbf{\boldmath$\Delta$}@5 \\
\midrule
\textbf{Learned (deployed)} & \textbf{0.459} & \textbf{0.534} & -- \\
\midrule
\multicolumn{4}{@{}l}{Layer 0 ($w_{\text{AFT}} < 1.6\times10^{-7}$)} \\
\quad Uniform ($w{=}0.5$) & 0.435 & 0.504 & $-$0.030 \\
\midrule
\multicolumn{4}{@{}l}{Layer 1 ($w_{\text{AFT}} = 0.584 \pm 0.063$)} \\
\quad Uniform ($w{=}0.5$) & 0.458 & 0.534 & 0.000 \\
\quad Mamba-only          & 0.437 & 0.512 & $-$0.022 \\
\quad AFT-only            & 0.430 & 0.495 & $-$0.039 \\
\midrule
\multicolumn{4}{@{}l}{Both layers} \\
\quad Uniform ($w{=}0.5$) & 0.427 & 0.495 & $-$0.039 \\
\bottomrule
\end{tabular*}
}

\vspace{1mm}
\begin{minipage}{\columnwidth}
\footnotesize
\textit{Note:} Computed over 1{,}915{,}476 keypoints across 648 HPatches images. The weights were trained jointly with the learned gate, so an override induces a distribution shift; these values therefore upper-bound the cost attributable to the gate itself and are reported as a diagnostic rather than as a retrained ablation.
\end{minipage}
\end{table}

\begin{table}[t]
\centering
\caption{Ablation on ORB~\cite{b1} using HPatches~\cite{b31}.}
\label{tab:ablation_hpatches}
\small
\renewcommand{\arraystretch}{1}
\resizebox{0.48\textwidth}{!}{%
\begin{tabular}{cccccc}
\toprule
\multicolumn{3}{c}{\textbf{Local Stage}} &
\multirow{2}{*}{\textbf{Context Stage}} &
\multicolumn{2}{c}{\textbf{ MMA} $\uparrow$} \\
\cmidrule(lr){1-3}\cmidrule(lr){5-6}
\textbf{Descriptor} & \textbf{MLP\textsubscript{geo}} & \textbf{LFF-MLP\textsubscript{geo}} &  & \textbf{@3} & \textbf{@5} \\
\midrule
\cmark &        &        &        & 0.403 & 0.448 \\
\cmark & \cmark &        &        & 0.427 & 0.488 \\
\cmark &        & \cmark &        & 0.431 & 0.491 \\  
\cmark & \cmark &        & \cmark & 0.448 & 0.519 \\   
\cmark &        & \cmark & \cmark & \textbf{0.459} & \textbf{0.534} \\
\bottomrule
\end{tabular}}
\end{table}


\begin{table}[t]
\centering
\renewcommand{\arraystretch}{1.1}
\caption{Comparison of spatial serialization strategies on HPatches~\cite{b31} with ORB keypoints. All settings were retrained under identical settings.}
\begin{tabular*}{\columnwidth}{@{}l@{\extracolsep{\fill}}cc@{}}
\toprule
\textbf{Serialization} & \textbf{MMA @ 3 $\uparrow$} & \textbf{MMA @ 5 $\uparrow$} \\
\midrule
Detector order (none) & 0.453 & 0.528  \\
Row-major scanning    & 0.441 & 0.507  \\
Hilbert curve         & 0.454   & 0.529  \\
Z-order (ours)        & \textbf{0.459} & \textbf{0.534} \\
\bottomrule
\end{tabular*}
\label{tab:serialization_ablation}
\end{table}

\begin{figure*}[t] 
    \centering
    \includegraphics[width=\textwidth]{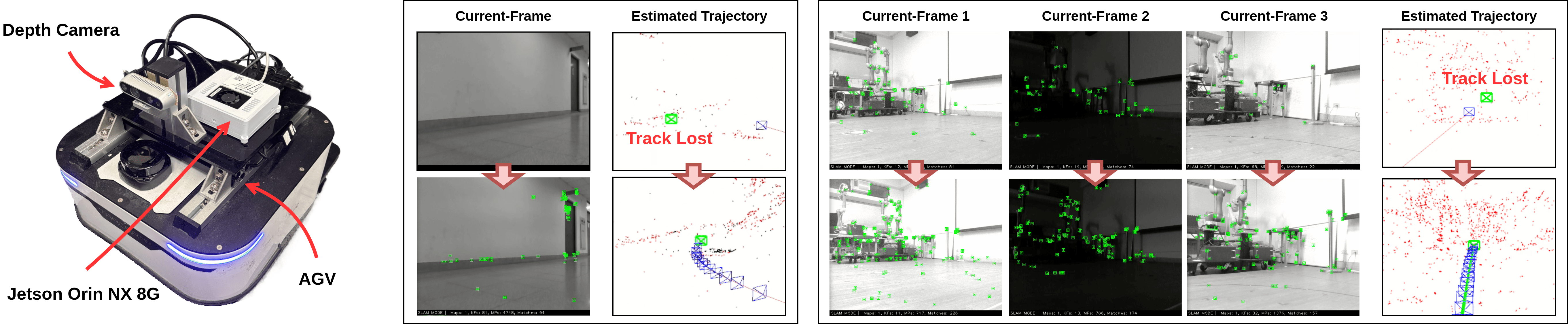} 
    \caption{Robot configuration (left) and qualitative tracking comparisons in a textureless corridor (middle) and a room with varying illumination (right). In each scenario, the top row shows the ORB baseline and the bottom row shows Desc++; Current-Frames 1--3 correspond to three moments under different illumination conditions.} 
    \label{fig:real} 
\end{figure*}

\subsection{Discussion}
\label{sec:discussion}
\subsubsection{Plug-and-play compatibility}
A design goal of Desc++ is compatibility with existing feature-based pipelines, and the system-level experiments provide evidence for this property. The same enhancement module was integrated, without any modification to the downstream tracking, mapping, or matching logic, into four heterogeneous systems: a stereo system (ORB-SLAM2~\cite{b2}), a stereo-inertial system (ORB-SLAM3~\cite{b3}), a visual-LiDAR system (RGB-L~\cite{b4}), and a multi-camera visual-inertial system (MAVIS-SLAM~\cite{b41}), improving trajectory accuracy on the majority of sequences relative to each base system (10 of 10 on EuRoC stereo, 9 of 11 on EuRoC stereo-inertial, 10 of 11 on KITTI, and 9 of 9 on Hilti). This portability contrasts with replacement-based front-ends: as shown in Table~\ref{tab:runtime_cost}, adopting a learned matching pipeline such as Rover-SLAM~\cite{b42} achieves higher accuracy on several sequences but requires restructuring the front-end and an order of magnitude more GPU memory, whereas Desc++ attains competitive accuracy while preserving the original architecture with only a marginal runtime overhead. Compatibility, however, also implies inheritance: by retaining the original extractor and matcher, Desc++ inherits their failure modes. The same design constraint applies to training: Desc++ optimizes descriptor matching rather than a system-specific trajectory objective, because the latter would couple the module to a particular SLAM
estimation pipeline and conflict with its plug-and-play deployment across heterogeneous systems. Geometric supervision is nevertheless incorporated during training-data construction through reprojection-based correspondence labels, making matching AP a direct objective for the correspondence quality that the module is designed to improve.

\subsubsection{Limitation}
Since Desc++ refines descriptors after feature extraction, its effectiveness is inherently bounded by the information content of the input features. When the extractor itself fails to detect meaningful patterns (as in the severe motion blur of EuRoC V203 under a purely visual configuration), no enhancement can recover the missing information, and tracking failure occurs regardless of the enhancer. In particular, the baseline and FeatureBooster~\cite{b7} also fail on this sequence, indicating a boundary of the descriptor-enhancement paradigm rather than a deficiency of the proposed module. This limitation can be mitigated at the system level: with inertial fusion, motion predictions bridge the blurred segments, allowing Desc++ to re-establish reliable associations once features stabilize (Table~\ref{tab:euroc}). The two real-world sequences in Fig.~\ref{fig:real} complement this boundary case from the other side. We recorded them on a custom mobile platform equipped with an Intel RealSense D435i camera (30\,Hz RGB-D, 200\,Hz IMU), running ORB-SLAM3 in RGB-D-inertial mode with all processing on an NVIDIA Jetson Orin NX (8\,GB). The first sequence is a long texture-less corridor with a 90$^\circ$ turn, and the second is a room with severe illumination changes. Unlike the motion-blur case, the features in these scenes remain detectable but lack discriminability, placing them within the operating regime of descriptor enhancement.

As no external ground-truth reference is available for these sequences, we do not evaluate trajectory accuracy on them. The comparison is qualitative and concerns tracking continuity only: the baseline lost tracking in both scenes, whereas the Desc++-enhanced system maintained continuous tracking. Continuous tracking is a necessary condition for, but not evidence of, an accurate trajectory, so these sequences illustrate the operating regime rather than quantify the benefit within it. Together, the motion-blur and real-world cases delineate the operating regime of descriptor enhancement: it helps when features remain detectable, but their descriptors are insufficiently discriminative, and it cannot compensate for extraction failures. Quantitative evaluation under nighttime, adverse-weather, and severe-motion-blur conditions on benchmarks with ground-truth trajectories remains future work.

Our evaluation also leaves several aspects open. First, although the keypoint-count analysis shows stable gains from 64 to 5000 keypoints, robustness to stronger domain shifts, such as cross-season or cross-weather appearance changes, remains unestablished. Since Desc++ operates on extracted descriptors and keypoint geometry rather than raw images, its behavior under such shifts is partly inherited from the underlying feature extractor and descriptor. Second, Desc++ is designed for sparse descriptor-based matching and therefore cannot be directly applied to purely dense photometric pipelines that lack this interface. Hybrid systems that retain sparse descriptor-based components could nevertheless incorporate the proposed enhancement at those components. Third, Desc++ enhances descriptors without explicitly estimating their reliability. A natural extension is therefore to predict a per-descriptor confidence alongside the enhanced representation, allowing the SLAM back-end to down-weight potentially unreliable associations; we leave this joint enhancement-and-confidence design to future work. Finally, while the current implementation achieves real-time descriptor enhancement on embedded hardware, it relies on PyTorch inference, and the complete Desc++-enhanced SLAM pipeline incurs additional computational overhead compared with the baseline. TensorRT deployment and INT8 quantization could reduce this overhead; the MLP components are directly supported, whereas the selective-scan operator requires a dedicated plugin. Such optimizations also require end-to-end accuracy and runtime validation, and are left for future work.


\section{Conclusion}
In this paper, we have presented Desc++, a lightweight descriptor enhancement module that improves visual data association in existing V-SLAM systems without modifying their feature extraction or matching pipelines. By combining sequential state-space modeling over Z-order serialized keypoints with attention-free global context aggregation in a parallel hybrid architecture, Desc++ captures richer contextual relationships than prior enhancement methods while retaining linear-time complexity. Experiments across descriptor matching, data association, and system-level benchmarks showed that the enhanced descriptors improve matching accuracy on HPatches, increase the number of tracked map points, and reduce trajectory errors across four heterogeneous SLAM systems on the EuRoC, KITTI, and Hilti benchmarks, with particularly strong gains in the multi-camera configurations involving substantially different viewpoints. Compared with a replacement-based learned front-end, Desc++ achieves competitive accuracy at an order of magnitude lower memory cost while sustaining real-time performance on embedded hardware. Our analysis further delineates the operating regime of descriptor enhancement: while it is bounded by failures at the extraction stage, it is effective whenever features are degraded while still detectable, making it a practical path to improving deployed real-time V-SLAM systems.


\section*{Acknowledgment}
This work was financially supported in part by the National Science and Technology Council of Taiwan and also by the Taiwan Centers of Excellence in Intelligent Team Robotics of the Ministry of Education of Taiwan.

\begin{IEEEbiography}[{\includegraphics[width=1in,height=1.25in,clip,keepaspectratio]{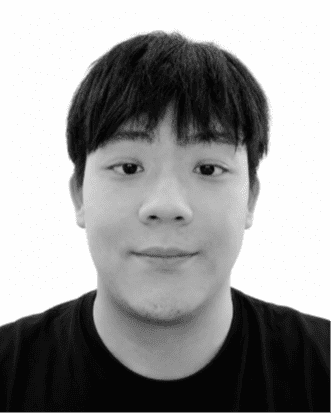}}]{Ting-Wei Ou} was born in Taipei, Taiwan, 2001. He received his B.S. degree in mechanical engineering from National Yang Ming Chiao Tung University, Taiwan, in 2023.  He is currently pursuing an M.S. degree in robotics at National Yang Ming Chiao Tung University, Taiwan. His research interests include robotic perception, visual SLAM, and resource-efficient perception for mobile robots.
\end{IEEEbiography}

\begin{IEEEbiography}[{\includegraphics[width=1in,height=1.25in,clip,keepaspectratio]{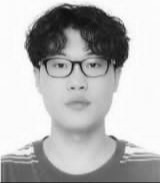}}]{Huang-Ting Lin} was born in Tainan, Taiwan, in 2002. He received his bachelor’s degree in mechanical engineering from National Taiwan University of Science and Technology, Taiwan, in 2025. He is currently pursuing an M.S. degree in electrical control engineering at National Yang Ming Chiao Tung University, Taiwan. His research interests include computer vision for robotics and visual SLAM.
\end{IEEEbiography}

\begin{IEEEbiography}[{\includegraphics[width=1in,height=1.25in,clip,keepaspectratio]{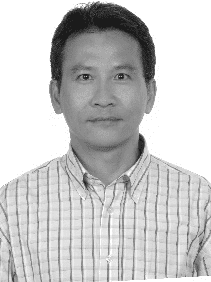}}]{Kuu-Young Young} (M’86–SM’04) was born in Kaohsiung, Taiwan, 1961. He received his B.S. degree in electrical engineering from National Taiwan University, Taiwan, in 1983, and M.S. and Ph.D. degree in electrical engineering from Northwestern University, Evanston, IL, U.S.A., in 1987 and 1990, respectively. Since 1990, he has been with the Department of Electrical Engineering at National Yang Ming Chiao Tung University, Hsinchu, Taiwan, where he is currently a Professor. He served as the chairman of the department from 2003 to 2006, and the associate dean of Electrical and Computer Engineering College, NYCU, from 2007 to 2010 and also 2014 to 2015. His research interests include robot compliance control, robot learning control, robot calibration and path planning, teleoperation, and assistive robots.
\end{IEEEbiography}

\end{document}